\pdfoutput=1
\documentclass[11pt]{article}

\usepackage[]{acl}

\usepackage{times}
\usepackage{latexsym}

\usepackage[T1]{fontenc}

\usepackage[utf8]{inputenc}

\usepackage{microtype}
\usepackage{microtype}
\usepackage{graphicx}
\usepackage{amsmath}
\usepackage{amsfonts}
\usepackage{amssymb, bbm}
\usepackage{booktabs}
\usepackage{multicol}
\usepackage{multirow}
\usepackage{ulem}
\usepackage{float}
\usepackage{url}
\usepackage{makecell}
\usepackage{subfig}
\usepackage{listings}
\usepackage{pifont}
\usepackage[T1]{fontenc}
\usepackage{makecell}
\usepackage{cellspace}
\usepackage{multirow}
\usepackage{balance}
\usepackage{graphicx}
\usepackage{utfsym}
\usepackage{diagbox}
\usepackage{colortbl}
\usepackage[most]{tcolorbox}
\usepackage{diagbox}
\usepackage{xcolor}
\definecolor{mygreen}{RGB}{209,255,200}
\definecolor{myred}{RGB}{255,205,196}
\definecolor{bys}{RGB}{0,112,192}
\usepackage{soul} 
\usepackage{color, xcolor} 

\usepackage{enumitem}
\usepackage{color}
\definecolor{Red}{RGB}{255,0,0}
\definecolor{Orange}{RGB}{237,125,49}
\definecolor{Green}{RGB}{0,128,0}
\definecolor{Blue}{RGB}{0,112,192}

\definecolor{bblue}{HTML}{4F81BD}
\definecolor{rred}{HTML}{C0504D}
\definecolor{ggreen}{HTML}{9BBB59}
\definecolor{ppurple}{HTML}{9F4C7C}
\definecolor{darkGreen}{rgb}{0.2,0.5,0.2}
\definecolor{mydarkblue}{rgb}{0,0.08,0.45}
\definecolor{mygray}{gray}{.9}
\colorlet{LightOrange}{myred!20!}

\definecolor{forestgreen}{RGB}{0,128,0}

\title{
\includegraphics[scale=0.012]{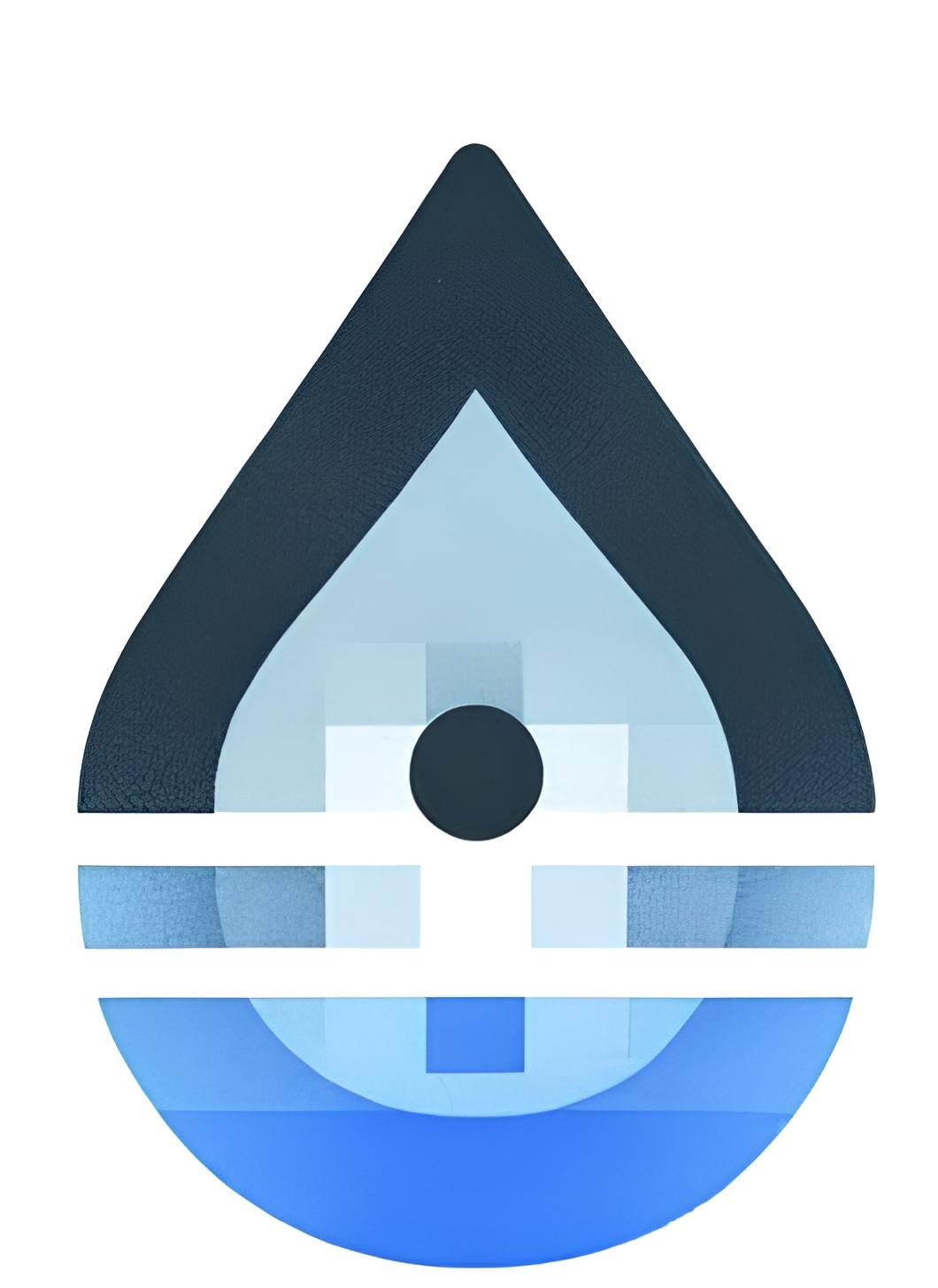} WaterBench: Towards Holistic Evaluation of Watermarks for \\ Large Language Models  
}

\author{Shangqing Tu$^1$\thanks{~~Equal Contribution.}, Yuliang Sun$^{2*}$, Yushi Bai$^1$, Jifan Yu$^1$,  Lei Hou$^1$\thanks{~~Corresponding author.} \and   Juanzi Li$^1$\\
 $^1$Department of Computer Science and Technology, Tsinghua University, Beijing 100084, China \\
 $^2$School of Computer Science and Engineering, Beihang University \\
  \texttt{\{tsq22,bys22,yujf21\}@mails.tsinghua.edu.cn} \\
  \texttt{21371245@buaa.edu.cn,\{houlei,lijuanzi\}@tsinghua.edu.cn}  \\
}

\begin{document}
\maketitle
\begin{abstract}
To mitigate the potential misuse of large language models (LLMs), recent research has developed watermarking algorithms, which restrict the generation process to leave an invisible trace for watermark detection. Due to the two-stage nature of the task, most studies evaluate the generation and detection separately, thereby  presenting a challenge in unbiased, thorough, and applicable evaluations.
In this paper, we introduce WaterBench, the first comprehensive benchmark for LLM watermarks, in which we design three crucial factors: (1) For \textbf{benchmarking procedure}, to ensure an apples-to-apples comparison, we first adjust each watermarking method's hyper-parameter to reach the same watermarking strength, then jointly evaluate their generation and detection performance. (2) For \textbf{task selection}, we diversify the input and output length to form a five-category taxonomy, covering $9$ tasks.  (3) For \textbf{evaluation metric}, we adopt the  GPT4-Judge for automatically evaluating the decline of instruction-following abilities after watermarking. We evaluate $4$ open-source watermarks on $2$ LLMs under $2$ watermarking strengths and observe the common struggles for current methods on maintaining the generation quality. The code and data are available at \url{https://github.com/THU-KEG/WaterBench}.

\end{abstract}

\section{Introduction}

\begin{figure}[t] 
\centering 
\includegraphics[width=0.95\linewidth]{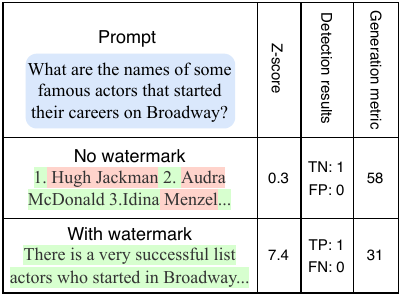} 
\caption{The generated texts without and with watermark \citep{kirchenbauer2023watermark} on a test example from AlpacaFarm~\cite{dubois2023alpacafarm}, an instruction-following benchmark. 
LLM equipped with watermark will be more inclined to generate tokens in the \colorbox{mygreen}{green list}, which can then be detected by a higher z-score measurement ($z>4$).
We utilize TP, TN, and GM to jointly evaluate the watermarking performance.}
\label{fig:case_table} 
\end{figure}

LLM has achieved significant success in generating human-like texts~\cite{cai2023does,openai2023gpt4,bubeck2023sparks}.  However, the potential misuse of LLM has also raised concerns~\cite{li2023multi}. For example, ChatGPT can be used to generate fake news~\cite{wang2023implementing}, which may manipulate the public opinion. To mitigate this kind of risk, it is necessary to develop a watermarking algorithm to detect whether a text is generated by LLM~\cite{kirchenbauer2023watermark}. As shown in Figure~\ref{fig:case_table}, the watermarked texts are generated with a biased distribution of tokens, which distinguishes it from unwatermarked texts. We believe the goal of watermarking is to achieve high detection accuracy while maintaining the generation quality. So we utilize the commonly used TP (True Positive), TN (True Negative), and GM (Generation Metric) to  evaluate watermarks~\cite{ghosal2023towards}. 

\begin{table*}[t]
  \centering
  \small
  \begin{tabular}{cccccc}
  \toprule
  \multicolumn{1}{c}{Research for Watermark}  & \begin{tabular}[c]{@{}c@{}}  Control\\  Hyper Para.\end{tabular} & \begin{tabular}[c]{@{}c@{}} Jointly Test \\ 
  (TP,TN,GM)  \end{tabular} & \begin{tabular}[c]{@{}c@{}} Number\\  of Tasks  \end{tabular} & \begin{tabular}[c]{@{}c@{}} Instruction\\ Following \end{tabular} & \begin{tabular}[c]{@{}c@{}}Metric for \\ Generated Text
  \end{tabular}  \\ 
  \midrule
  LM Watermark~\cite{kirchenbauer2023watermark}    & $\checkmark$  & $\times$    & 1     & $\times$   & Perplexity \\
  V2 Watermark~\cite{kirchenbauer2023reliability}      &  $\checkmark$ & $\times$   & 2    & $\times$  & Perplexity  \\
  Robust Watermark~\cite{kuditipudi2023robust}     & $\checkmark$ & $\times$    & 1    & $\times$   & Perplexity \\
  
  GPT Watermark~\cite{zhao2023provable}   & $\times$ & $\times$    & 2    & 
   $\times$ & Perplexity   \\
  Semantic Watermark~\cite{fu2023watermarking}     & $\checkmark$ & $\times$     & 2     & $\times$ & Ref.  \\
  Three Bricks~\cite{fernandez2023three}   & $\checkmark$   & $\times$    & 3    &  $\times$ & Ref.  \\
  \midrule
  \textbf{WaterBench (ours)}  & $\checkmark$ & $\checkmark$   & 9  &  $\checkmark$ & Ref./GPT4-Judge  \\ 
  \bottomrule
  \end{tabular}
  \caption{Comparison with existing works' evaluations of LLM watermarks. The column \textit{Jointly Test} means whether the three metrics for each watermark are jointly tested under one run. The column \textit{Instruction Following} means whether it evaluates this ability. The term \textit{Para.} and  \textit{Ref.} are short for parameter and reference-based metric.}
   \label{tab:comp}
  \end{table*}

Due to the two-stage nature of this task, most studies~\cite{kuditipudi2023robust,zhao2023provable} evaluate the generation and detection separately and they do not conduct a unified hyper-parameter search for each watermarking method, which may lead to unfair comparisons. Since, there is usually a trade-off between the detection performance and the generation quality. Besides, previous evaluations are often conducted via text completion on a single dataset, such as C4 RealNewsLike dataset~\cite{raffel2020exploring}, which cannot comprehensively measure the generation quality of LLMs.  Furthermore, most evaluations only calculate the perplexity~\cite{kirchenbauer2023reliability}, which is not aligned with human preference and thus not practical in the era of LLMs~\cite{chia2023instructeval}.

To address these issues, we propose WaterBench, the first comprehensive benchmark for LLM watermarks, which has three crucial factors: (1) \textbf{Benchmarking Procedure}: We first introduce the concept of watermarking strength~\cite{mei2002decision}, i.e. the detection robustness to disturbance, to quantify the LLM watermarks' trade-off controlled by hyper-parameters. We present a reasonable hyper-parameter search procedure: Given a dataset and an LLM, we adjust the hyper-parameters of each watermarking method to unify the watermarking strength and then freeze the parameters to jointly evaluate the detection and generation performance. (2) \textbf{Task Selection}: To add disturbance on watermarks, we differentiate the task settings based on the length of input and output, which decides how much information the watermark can embed. Therefore, we form a new taxonomy with five task categories and nine sub-tasks, which are selected from existing datasets with various length settings~\cite{dubois2023alpacafarm}.
(3) \textbf{Evaluation Metric}: We adopt the GPT4-Judge~\cite{zheng2023judging} for automatically evaluating the instruction-following performance decline after watermarking. Then we conduct a human evaluation to verify the agreement between the human and GPT4.


Based on the WaterBench dataset, we conduct an experiment of 4 reproducible watermarks on 2 LLMs (Llama2-chat~\cite{touvron2023llama} and InternLM~\cite{2023internlm}), leading to some interesting findings: (1)  We adjust two different watermarking strengths, 0.7 and 0.95, and observe that the detection and generation performance are significantly different. In other words, if we compare two watermark strategies without aligning their watermarking strengths, it is easy to let one ``surpass'' another in some aspects. (2) The tasks with short output length are generally more difficult to detect, with lower TP. The V2 watermark~\cite{kirchenbauer2023reliability} is the best watermarking method in terms of GM.  (3) On the open-ended task, if we use GPT4-judge to evaluate, the watermarked LLM will decrease over $96\%$ from original LLM, which shows the sensitivity of the metric and indicates the common struggles of watermarks on maintaining the generation quality. In human evaluation, the GPT4 obtains over 0.6 Cohen's kappa coefficient with $3$ annotators, achieving substantial agreement. 


To summarize, our contributions are three-fold: (1) We propose a new benchmarking procedure that first search hyper-parameters for watermarks then jointly evaluate detection and generation performance to eliminate the unfair comparison between different watermarking strengths. (2) We construct a multi-task benchmark to facilitate future research. (3) We incorporate GPT4-Judge to evaluate the watermarked LLMs, which effectively reflects the decline of generation quality.

\section{Related Work}

To detect LLM-generated text, previous works~\cite{tu2023chatlog,guo2023close,mitchell2023detectgpt} mainly explored the classifiers that distinguishes human and LLM-generated texts based on features. However, as LLMs are becoming more and more alike human, some classifiers may mistakenly recognize human as LLMs~\cite{sadasivan2023can}.

\begin{figure*}[t] 
  \centering 
  \includegraphics[width=\linewidth]{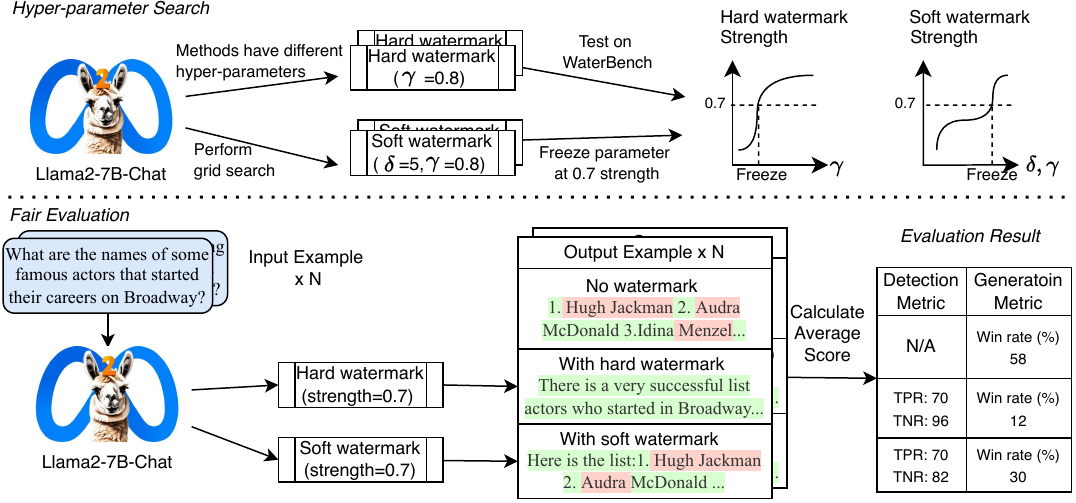}
  \caption{An illustration of the evaluation process on WaterBench. Given an LLM, a watermarking method and our benchmark, we first search the hyper-parameter to fix the watermarking strength of each method, then jointly evaluate their detection and generation performance for fair comparisons. }
  \label{fig:motivation_example} 
\end{figure*}
In addition to black-box classifiers, recent approaches have also introduced white-box detection methods that inject watermarks into LLM generated texts~\cite{tang2023science,yang2023watermarking,liu2024survey}. The inference-time watermarks~\cite{pan2024markllm} randomly split the vocabulary and adjust the probability distribution at each decoding step, which guarantees the presence of detectable patterns, known as watermarks, in the generated text. Some works~\cite{kirchenbauer2023reliability,liu2023semantic} focus on improving the detection robustness to paraphrasing attacks~\cite{krishna2023paraphrasing} or at low-entropy environment~\cite{lu2024entropybased}. Other works like unbiased watermark~\cite{hu2023unbiased} and NS-watermark~\cite{takezawa2023necessary} focus on improving the quality of generated texts~\cite{hou2024ksemstamp,li2023improving}. 

On the other hand, post-hoc watermark~\cite{atallah2001natural,topkara2005natural} is also a line of research, which insert watermarks into texts by synonym replacement~\cite{yang2023watermarking,yoo2023robust} or paraphrasing~\cite{munyer2023deeptextmark}. Recently, \citet{sato2023embarrassingly} presented a simple but effective method that replace each space character with another codepoint of whitespace. However, this simple watermark can also be easily erased.

\section{WaterBench}
\label{sec:WaterBench}

To investigate how inference-time watermarks perform on  detection and generation, as shown in Figure ~\ref{fig:motivation_example}, we propose a benchmarking procedure  that ensures fair comparisons (Section ~\ref{sec:benchmarking_procedure}). Then, we present the WaterBench dataset with a diverse length distribution (Section ~\ref{sec:data_collection}). Finally, we introduce the  GPT4-Judge evaluation (Section ~\ref{sec:evaluation_metric}).

\subsection{Problem Definition for Watermarking}

\paragraph{Generation Stage} Assume an auto-regressive LLM  $\theta$ has a vocabulary $V$, the probability distribution for the $t$-th token in a sequence $\mathbf{S}=\{s_1, s_2, \dots, s_{|\mathbf{S}|}\}$ can be expressed as:

\begin{equation}
p(s_{t}) = p_{\theta}(s_t | s_{<t})
\label{lms}
\end{equation}

LLM predicts the $p(s_{t})$ by calculating a logit vector $l^{(t)} \in \mathbb{R}^{|V|}$ for each item $k$ in vocabulary. \citet{kirchenbauer2023watermark} propose $2$ watermarks, namely hard and soft watermarks, to add watermarks to text by imposing restrictions on the vocabulary during each decoding step. Specifically, the ``Hard Red List'' watermark algorithm randomly divides the vocabulary into ``green'' and ``red'' lists using a hash function. During the generation process, only tokens from the green list can be chosen for the $t$-th position.  While soft watermark approach introduces a constant $\delta$ to the logit $l^{(t)}_k$ of tokens in the green list during the prediction step:

\begin{equation}
p^{(t)}_k = \exp (l^{(t)}_k + \delta) / \sum_{i} \exp (l^{(t)}_i)
\end{equation}

\begin{table*}[t]
  \centering  
  \scriptsize
  \resizebox{\textwidth}{!}{
  \begin{tabular}{lclrccc}
  \toprule
  \textbf{Category \& Source Data} & \textbf{ID} & \textbf{Task} & \textbf{Metric} & \textbf{Language} & \textbf{\#data} & \textbf{Len.Input / Answer} \\
  \midrule
  \emph{(Short Input, Short Answer)} \\
  KoLA~\cite{yu2023kola} & 1-1 & Entity Probing & F1 & English & 200 & 7.72 / 2.96 \\
  Copen~\cite{peng2022copen} & 1-2 & Concept Probing & F1 & English & 200 & 51.52 /  1.57 \\
  
  \midrule
  \emph{(Short Input, Long Answer)} \\
  ELI5~\cite{fan-etal-2019-eli5} & 2-1 & Long-form QA & Rouge-L & English & 200 & 41.04 / 236.6 \\
  FiQA~\cite{maia201818} & 2-2 & Finance QA & Rouge-L & English & 200 & 13.67 / 251.13 \\
  \midrule
  \emph{(Long Input, Short Answer)} \\
  HotpotQA~\cite{yang2018hotpotqa} & 3-1 & Multi-Doc QA & F1 & English & 200 & 10619.4 / 2.65 \\
  LCC~\cite{chen2021evaluating} & 3-2 & Code Completion & Edit Sim & Python/C\#/Java & 200 & 2263.32 / 9.45 \\
  
  \midrule
  \emph{(Long Input, Long Answer)} \\
  MultiNews~\cite{fabbri2019multi} & 4-1 & Multi-Doc Summ. & Rouge-L & English & 200 & 2198.65 / 260.88 \\
  QMsum~\cite{zhong2021qmsum} & 4-2 & Query-Based Summ.  & Rouge-L & English & 200 & 12457.93 / 76.52 \\
  \midrule
  \emph{(Open Ended Generation)} \\
  AlpacaFarm~\cite{dubois2023alpacafarm} & 5-1 & Instruction Following & GPT4-Judge & English & 805 & 32.58 / 64.13 \\
  \bottomrule
  \end{tabular}
  }
  \caption{An overview of the dataset statistics in WaterBench.  `Dataset' denotes the origin of the sub-dataset. `Len.Input / Answer' refer to the average length of input question and reference answer.}
  \label{tab:stat}
  \end{table*}

\paragraph{Detection Stage}  To detect the presence of the watermark in the generated text, statistical analysis techniques such as the \emph{one proportion z-test} can be applied. While the hash function generates the green lists with a greenlist fraction $\gamma$, we can extract the watermark by re-computing the greenlist at each position to get a set of greenlist tokens $\mathbf{S}_g$. Then the significance can be derived by $z$-score:
\begin{equation}
z =  (|\mathbf{S}_g|- \gamma |\mathbf{S}| ) / \sqrt{\gamma (1-\gamma) |\mathbf{S}|}
\end{equation}

If the $z$-score is above the threshold, which means the corresponding P-value is small, then we can ensure that the text $\mathbf{S}$ is watermarked.


\subsection{Benchmarking Procedure}  
\label{sec:benchmarking_procedure}

Due to the two-stage nature of the task, previous works may use different hyper-parameters when testing the detection and generation, leading to unfair comparisons. 
As shown in Figure~\ref{fig:motivation_example}, we propose a fair benchmarking procedure to jointly evaluate the detection and generation performance.


\paragraph{Watermarking Strength.} 
To retain consistency with the two stages and maintain fairness in our evaluations, we introduce the concept of watermarking strength~\cite{mei2002decision} into LLM watermarks. In image watermarking~\cite{akhaee2009robust}, the higher watermarking strength means the better robustness for detecting the watermark.  In LLM watermarking, we believe the watermarking strength should be independent with the referenced answer and can measure the detecting robustness. Therefore, we define the watermarking strength as the ratio of the number of watermarked texts that are correctly detected to the total number of watermarked texts, namely the TPR (True Positive Rate), which is a definite value after setting the input, the watermarking algorithm and its hyper-parameters. By freezing the watermarking strength, the evaluation results can remain consistent in the two stages. Some methods may adaptively set their strength for each case~\cite{takezawa2023necessary} and that's not discussed in this paper yet we leave for the future.

\paragraph{Hyper-Parameter Search.} 
Although the watermarking strength depends on hyper-parameters, the hyper-parameters of different watermarking methods are not comparable~\cite{ghosal2023towards}. Therefore, we propose a hyper-parameter search procedure to unify the watermarking strength of different watermarking methods. Specifically, we first set the hyper-parameters of each watermarking method to the initial value by default, then we perform grid search~\cite{alibrahim2021hyperparameter} to change the watermarking strength to the desired level and minimize the changes to hyper-parameters. Finally, we fix the hyper-parameters to the determined values and jointly evaluate the two-stage performance.

For researchers aiming to introduce a new watermark candidate to WaterBench, a suitable hyper-parameter should first be identified by them to achieve a certain True Positive Rate (TPR), for instance, 0.95. Subsequently, the evaluation code can be executed to obtain True Negative (TN) and Generation Metric (GM) results. Ultimately, they can benchmark their performance against other watermarks with an equivalent TPR on our benchmark.

\subsection{Task Selection}
\label{sec:data_collection}

As shown in Table ~\ref{tab:stat}, we select nine typical tasks for five distinct task settings, covering a wide range of input and output length, including:

\begin{figure*}[t]
  \centering
  \includegraphics[width=\linewidth]{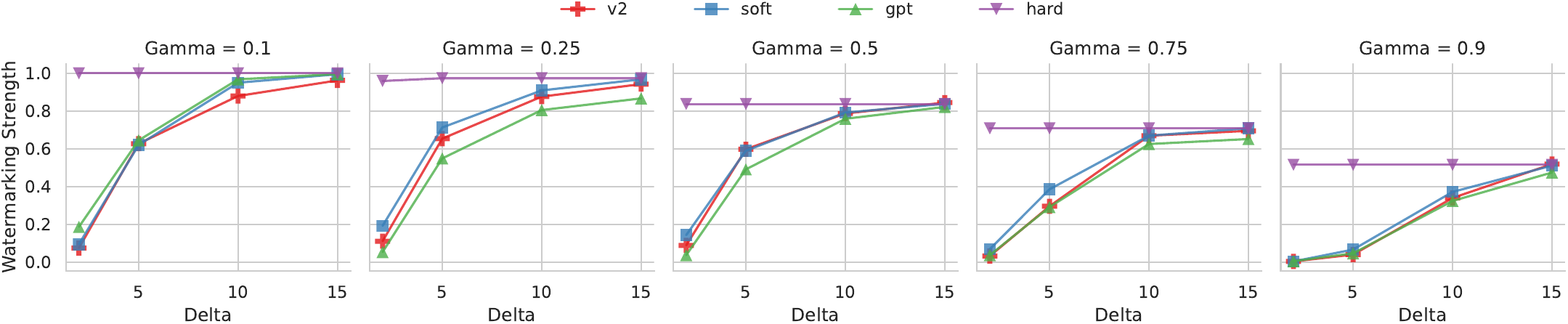}
  \caption{ 
  The watermarking strength results of $4$ watermarking methods on Llama2-7B-chat after the hyper-parameter search for $\delta$ and $\gamma$. The watermarking strength is measured by the average TPR on our WaterBench.
  }
  \label{fig:hyper_res}
\end{figure*}

\paragraph{Category 1: Short Input, Short Answer.}  As the input and answer length decides how much information that the watermarking algorithm can hide, we first choose two tasks that have short input and answer length to disturb the watermarking methods. Both tasks evaluate the \textit{Factual knowledge} in a close-ended setting. The task 1-1 is the knowledge probing, we use $200$ triplets from KoLA dataset~\cite{yu2023kola} with different frequency in Wikipedia to probe the facts from LLMs. For task 1-2, the concept probing, we use the $200$ samples from the cic and csj task in Copen dataset~\cite{peng2022copen}. As the output length is short, we use the F1 score as the evaluation metric.

\paragraph{Category 2: Short Input, Long Answer.} To control 
for the variable of answer length, we choose another $2$ tasks with the short input but long answer. Both tasks belong to \textit{Long-form QA}, which is the common format that users interact with LLMs, where user ask a short question and expect a long answer. For task 2-1, we use $200$ samples from the ELI5 dataset, which is a long-form question-answering dataset composed of threads from the Reddit forum "Explain Like I'm five". For long-form QA with finance knowledge (Task 2-2), we use $200$ samples from the FiQA dataset.

\paragraph{Category 3: Long Input, Short Answer.}   To control the variable of input length, we select  $2$ tasks from LongBench~\cite{bai2023longbench} with long input and short output. To evaluate the effect of watermarks of LLMs on the \textit{reasoning} (Task 3-1), we select  $200$ samples from the HotpotQA dataset~\cite{yang2018hotpotqa}, which is a multi-hop question-answering dataset. For \textit{code completion} (Task 3-2), we use $200$ samples from the LCC dataset~\cite{chen2021evaluating}, a dataset constructed by filtering code within a single file from GitHub.

\paragraph{Category 4: Long Input, Long Answer.} To control both input and output length, we involve $2$ tasks with long inputs and answers. The $2$ tasks are both \textit{Summarization} task, which is a particular skill for serving people's information needs. We select $200$ samples from  the widely-used multi-document news summarization dataset, MultiNews~\cite{fabbri2019multi}. For query-based summarization, we use $200$ samples from the QMSum dataset~\cite{zhong2021qmsum} with both input documents and queries for specific parts of the documents.

\paragraph{Category 5: Open-Ended Generation.}  While the aforesaid datasets mainly evaluate the specific skills of LLMs, the input and output length may be limited to a certain range that is suited for the corresponding task. In real world application of LLMs, the abilities to follow the user's instructions are also important where the generation is often open-ended. To comprehensively evaluate the \textit{instruction-following} performance of watermarked LLMs, we select the AlpacaFarm dataset~\cite{dubois2023alpacafarm}, which contains $805$ instructions, consisting of 5 different sources of instructions, with $32.58$ tokens in the input and $64.13$ tokens in the reference answer on average.

\subsection{Evaluation Metric}
\label{sec:evaluation_metric}

We used an evaluation method called GPT4-Judge~\cite{zheng2023judging} to compare how well watermarked LLMs and Davinci-003 could generate text on the AlpacaFarm dataset. The GPT4-Judge measures which model's output the GPT-4 system prefers when shown two responses for the same instruction. To be fair, the order of the models' output texts are randomly mixed up before GPT-4 chose to avoid the position bias~\cite{wang2023large}.

\section{Experiments}
\label{sec:Evaluation}

\begin{table*}[t]
  \centering  
  \resizebox{\textwidth}{!}{
  \begin{tabular}{l|cccc|cccc|cccc}
  \toprule
  \multirow{3}{*}{\textbf{Model}} & \multicolumn{4}{c|}{\textbf{C1: (Short Q, Short A)}} & \multicolumn{4}{c|}{\textbf{C2: (Short Q, Long A)}} &  \multicolumn{4}{c}{\textbf{C3: (Long Q, Short A)}}  \\
  & \multicolumn{4}{c|}{\textit{Factual Knowledge}} & \multicolumn{4}{c|}{\textit{Long-form QA} } & \multicolumn{4}{c}{\textit{Reasoning \& Coding}} \\
  & \textbf{TP} & \textbf{TN} & \textbf{GM} & \textbf{Drop} & \textbf{TP} & \textbf{TN} & \textbf{GM} & \textbf{Drop} & \textbf{TP} & \textbf{TN} & \textbf{GM} & \textbf{Drop} \\
  \midrule
  \rowcolor{mygray} Llama2-7B-chat & -- & -- & 17.8 & -- & -- & -- & 21.3 & -- & -- & -- & 37.5 & --\\
  + hard watermark & 89.5 & 100.0 & 5.0 & $\downarrow$ 71.9\% & 99.8 & 99.8 & 12.1 & $\downarrow$ 43.4\% & 82.5 & 100.0 & 16.4 & $\downarrow$ 56.3\%\\
  + soft watermark & 90.2 & 98.8 & 7.7 & $\downarrow$ 56.6\% & 100.0 & 100.0 & 9.9 & $\downarrow$ 53.3\% & 80.2 & 100.0 & 19.8 & $\downarrow$ 47.2\%\\
  + gpt watermark & 95.8 & 100.0 & 13.6 & $\downarrow$ 23.9\% & 100.0 & 92.5 & 5.2 & $\downarrow$ 75.5\% & 85.0 & 98.9 & 14.7 & $\downarrow$ 60.7\%\\
  + v2 watermark & 83.8 & 100.0 & 11.2 & $\downarrow$ 37.1\% & 100.0 & 100.0 & 13.3 & $\downarrow$ 37.4\% & 81.0 & 100.0 & 13.9 & $\downarrow$ 63.0\%\\
  
  \rowcolor{mygray} Internlm-7B-8k & -- & -- & 26.3 & -- & 12.2 & -- & 18.4 & -- & -- & -- & 31.6 & --\\
  + hard watermark & 88.2 & 99.6 & 1.8 & $\downarrow$ 93.1\% & 96.0 & 99.8 & 9.6 & $\downarrow$ 47.9\% & 88.5 & 99.7 & 11.7 & $\downarrow$ 63.0\%\\
  + soft watermark & 80.7 & 99.6 & 6.2 & $\downarrow$ 76.3\% & 99.7 & 99.8 & 7.6 & $\downarrow$ 58.7\% & 89.8 & 99.7 & 10.6 & $\downarrow$ 66.5\%\\
  + gpt watermark & 97.8 & 100.0 & 3.2 & $\downarrow$ 87.8\% & 100.0 & 100.0 & 7.8 & $\downarrow$ 57.6\% & 86.0 & 100.0 & 11.4 & $\downarrow$ 63.8\%\\
  + v2 watermark & 85.6 & 100.0 & 11.0 & $\downarrow$ 58.4\% & 98.0 & 100.0 & 7.6 & $\downarrow$ 58.4\% & 92.6 & 100.0 & 15.8 & $\downarrow$ 50.1\%\\
  
  \bottomrule
  \end{tabular}
  }
  \caption{True Positive Rate (TP), True Negative Rate (TN), Generation Metric (GM) and Generation Quality Drop (Drop) for category 1, 2 and 3 tasks at the watermarking strength level of 0.95 with $z$-score threshold of 4. }
  \label{tab:level12}
  \end{table*}

  \begin{table*}[t]
  \centering 
  \resizebox{\textwidth}{!}{
  \begin{tabular}{l|cccc|cccc|cccc}
  \toprule
  \multirow{3}{*}{\textbf{Model}} & \multicolumn{4}{c|}{\textbf{C4: (Long Q, Long A)}} & \multicolumn{4}{c|}{\textbf{C5: Open-Ended}} &  \multicolumn{4}{c}{\textbf{Overall: (12345)}}  \\
  & \multicolumn{4}{c|}{\textit{Summarization}} & \multicolumn{4}{c|}{\textit{Instruction Following} } & \multicolumn{4}{c}{\textit{Detection
  \& Generation }} \\
  & \textbf{TP} & \textbf{TN} & \textbf{GM} & \textbf{Drop} & \textbf{TP} & \textbf{TN} & \textbf{GM} & \textbf{Drop} & \textbf{TP} & \textbf{TN} & \textbf{GM} & \textbf{Drop} \\
  \midrule
  \rowcolor{mygray} Llama2-7B-chat & -- & -- & 23.3 & -- & -- & -- & 54.7 & -- & -- & -- & 28.3 & --\\
  + hard watermark & 100.0 & 100.0 & 11.6 & $\downarrow$ 50.0\% & 100.0 & 98.8 & 1.1 & $\downarrow$ 98.0\% & 95.3 & 99.5 & 10.1 & $\downarrow$ 64.1\%\\
  + soft watermark & 100.0 & 100.0 & 10.2 & $\downarrow$ 56.3\% & 99.4 & 98.9 & 0.6 & $\downarrow$ 98.9\% & 94.9 & 99.5 & 10.7 & $\downarrow$ 62.3\%\\
  + gpt watermark & 100.0 & 99.8 & 7.2 & $\downarrow$ 69.1\% & 99.6 & 95.8 & 0.2 & $\downarrow$ 99.5\% & 96.7 & 96.9 & 9.1 & $\downarrow$ 67.9\%\\
  + v2 watermark & 100.0 & 99.8 & 11.6 & $\downarrow$ 50.2\% & 100.0 & 99.9 & 0.9 & $\downarrow$ 98.4\% & 94.1 & 99.9 & 11.2 & $\downarrow$ 60.3\%\\
  
  \rowcolor{mygray} Internlm-7B-8k & -- & -- & 17.8 & -- & -- & -- & 21.5 & -- & -- & -- & 23.3 & --\\
  + hard watermark & 96.0 & 100.0 & 6.4 & $\downarrow$ 64.3\% & 96.5 & 99.6 & 0.8 & $\downarrow$ 96.5\% & 93.7 & 99.7 & 6.6 & $\downarrow$ 71.6\%\\
  + soft watermark & 98.7 & 99.8 & 4.6 & $\downarrow$ 74.0\% & 97.4 & 99.5 & 0.3 & $\downarrow$ 98.6\% & 94.0 & 99.6 & 6.5 & $\downarrow$ 72.2\%\\
  + gpt watermark & 97.2 & 100.0 & 5.2 & $\downarrow$ 70.9\% & 99.3 & 99.4 & 0.5 & $\downarrow$ 97.7\% & 96.8 & 99.8 & 6.2 & $\downarrow$ 73.4\%\\
  + v2 watermark & 97.2 & 100.0 & 5.5 & $\downarrow$ 69.2\% & 97.7 & 99.4 & 0.5 & $\downarrow$ 97.7\% & 95.1 & 99.8 & 8.9 & $\downarrow$ 61.7\%\\
  \bottomrule
  \end{tabular}
  }
  \caption{True Positive Rate (TP), True Negative Rate (TN), Generation Metric (GM) and Generation Quality Drop (Drop) for category 4, 5 and all tasks at the watermarking strength level of 0.95 with $z$-score threshold of 4.}
  \label{tab:level345}
  \end{table*}

\subsection{Experimental Settings}
We choose $2$ popular LLMs as our baselines: Llama2-7B-chat~\cite{touvron2023llama} and Internlm-7B-8k~\cite{2023internlm}, both models are instruction-tuned to align with human preference.
We evaluate $4$ different representative watermarks on these $2$ LLMs on our WaterBench, including:
\begin{itemize}[itemsep=0pt, leftmargin=*]
    \item \textbf{Hard Watermark}: The hard watermark~\cite{kirchenbauer2023watermark} is a binary watermark that restricts the vocabulary of the model to a subset of words during decoding. 
    \item \textbf{Soft Watermark}: The soft watermark~\cite{kirchenbauer2023watermark} is a continuous watermark that divide $\gamma$ vocabulary and adds a constant $\delta$ on logits to encourage watermarked vocabularies. 
    \item \textbf{GPT Watermark}: The GPT watermark~\cite{zhao2023provable} simplifies the watermarking process with a fixed group of restricted vocabularies to improve robustness against editing  attacks. 
    \item \textbf{V2 Watermark}: The V2 watermark~\cite{kirchenbauer2023reliability} improves the soft watermark with different hashing schemes, including the LeftHash and SelfHash to secure better robustness to the paraphrasing attack.
\end{itemize}

As the evaluation procedure described in Section~\ref{sec:benchmarking_procedure}, we first adjust the watermarking strength of each watermark by grid search. As shown in Figure~\ref{fig:hyper_res}, we find that the watermarking strength of each watermark increases when $\delta$ increases and increases when $\gamma$ reduces. We then choose the watermarking strength of $0.95$ and $0.7$ for each watermark and freeze their hyper-parameters for further detection and generation evaluation. Apart from the grid search results, we also display the ROC curve of watermarks in Appendix ~\ref{sec:search_ruc_detail}.

\begin{table*}[t]
  \centering  
  \resizebox{\textwidth}{!}{
  \begin{tabular}{l|cccc|cccc|cccc}
  \toprule
  \multirow{3}{*}{\textbf{Model}} & \multicolumn{4}{c|}{\textbf{C1: (Short Q, Short A)}} & \multicolumn{4}{c|}{\textbf{C2: (Short Q, Long A)}} &  \multicolumn{4}{c}{\textbf{C3: (Long Q, Short A)}}  \\
  & \multicolumn{4}{c|}{\textit{Factual Knowledge}} & \multicolumn{4}{c|}{\textit{Long-form QA} } & \multicolumn{4}{c}{\textit{Reasoning \& Coding}} \\
  & \textbf{TP} & \textbf{TN} & \textbf{GM} & \textbf{Drop} & \textbf{TP} & \textbf{TN} & \textbf{GM} & \textbf{Drop} & \textbf{TP} & \textbf{TN} & \textbf{GM} & \textbf{Drop} \\
  \midrule
  \rowcolor{mygray} Llama2-7B-chat & -- & -- & 17.8 & -- & -- & -- & 21.3 & -- & -- & -- & 37.5 & --\\
+ hard watermark & 0.0 & 100.0 & 13.7 & $\downarrow$ 23.3\% & 100.0 & 100.0 & 19.4 & $\downarrow$ 8.9\% & 39.2 & 100.0 & 21.0 & $\downarrow$ 44.1\%\\
+ soft watermark & 0.0 & 100.0 & 13.8 & $\downarrow$ 22.6\% & 100.0 & 100.0 & 19.4 & $\downarrow$ 8.7\% & 41.2 & 100.0 & 20.6 & $\downarrow$ 45.1\%\\
+ gpt watermark & 11.8 & 100.0 & 17.0 & $\downarrow$ 4.4\% & 99.5 & 99.8 & 13.8 & $\downarrow$ 35.0\% & 25.1 & 100.0 & 17.3 & $\downarrow$ 53.9\%\\
+ v2 watermark & 0.0 & 100.0 & 14.9 & $\downarrow$ 16.6\% & 99.5 & 100.0 & 19.4 & $\downarrow$ 8.8\% & 39.8 & 100.0 & 25.1 & $\downarrow$ 33.2\%\\
  \midrule
 
  \multirow{3}{*}{\textbf{Model}} & \multicolumn{4}{c|}{\textbf{C4: (Long Q, Long A)}} & \multicolumn{4}{c|}{\textbf{C5: Open-Ended}} &  \multicolumn{4}{c}{\textbf{Overall: (12345)}}  \\
  & \multicolumn{4}{c|}{\textit{Summarization}} & \multicolumn{4}{c|}{\textit{Instruction Following} } & \multicolumn{4}{c}{\textit{Detection
  \& Generation }} \\
  & \textbf{TP} & \textbf{TN} & \textbf{GM} & \textbf{Drop} & \textbf{TP} & \textbf{TN} & \textbf{GM} & \textbf{Drop} & \textbf{TP} & \textbf{TN} & \textbf{GM} & \textbf{Drop} \\
  \midrule
   \rowcolor{mygray} Llama2-7B-chat & -- & -- & 23.3 & -- & -- & -- & 54.7 & -- & -- & -- & 28.3 & --\\
+ hard watermark & 91.8 & 100.0 & 19.9 & $\downarrow$ 14.4\% & 96.5 & 99.8 & 17.3 & $\downarrow$ 68.4\% & 70.7 & 99.9 & 18.4 & $\downarrow$ 35.1\%\\
+ soft watermark & 92.0 & 100.0 & 20.2 & $\downarrow$ 13.3\% & 95.4 & 99.8 & 19.0 & $\downarrow$ 65.2\% & 70.7 & 99.9 & 18.6 & $\downarrow$ 34.4\%\\
+ gpt watermark & 96.0 & 100.0 & 15.0 & $\downarrow$ 35.4\% & 93.4 & 99.9 & 4.1 & $\downarrow$ 92.5\% & 69.9 & 99.9 & 14.5 & $\downarrow$ 48.7\%\\
+ v2 watermark & 88.8 & 100.0 & 19.7 & $\downarrow$ 15.3\% & 94.0 & 99.9 & 17.0 & $\downarrow$ 68.9\% & 69.4 & 100.0 & 19.5 & $\downarrow$ 31.2\%\\
  
\bottomrule
  \end{tabular}
  }
  \caption{ True Positive Rate (TP), True Negative Rate (TN), Generation Metric (GM) and Generation Quality Drop (Drop) for all tasks at the watermarking strength level of 0.7 with $z$-score threshold of 4 for Llama2-7B-chat.}
  \label{tab:level345_strength2}
  \end{table*}

\subsection{Main Results}
\label{sec:Main_EXP}
We conduct evaluations at the 0.95 watermarking strength on each task and report watermarks' results in Table~\ref{tab:level12} and~\ref{tab:level345}. Here are our findings:

\paragraph{Detection Performance.} Among all tasks, the detection performance for the short answer tasks (Category 1 and 3) are significantly worse than other tasks. This is because that watermarked LLMs produce short responses for these tasks, which can not contain enough green-words for detection, resulting in low $z$-scores, making the  detectors more likely to fail on discovering the watermark.

For the overall detection performance, most watermarks can achieve high TP rates of around 95\% which is consistent with the fixed watermarking strength, while the TN rates are nearly 100\% for all methods. This indicates that current LLM watermarks~\cite{kirchenbauer2023watermark} are generally good at detecting watermarked texts while remaining a clear distinction from unwatermarked texts.


\paragraph{Generation Performance.}

However, in terms of generation quality, all watermarks lead to significant drops in GM compared to the original models. The hard watermark exhibits the largest decreases of over 50\% in most cases. The generation performance also declines more severely for the open-ended task with over 90\% drop. These findings suggest that current watermarks encounter challenges in maintaining the generation quality, particularly for instruction-following tasks.

Among different watermarks, as shown in Table ~\ref{tab:level345}, the V2 watermark achieves higher GMs in most task categories, highlighting its effectiveness in preserving generation quality. The soft watermark and GPT watermark also exhibit competitive performance. While V2 watermark even shows better True Negative rate in most categories, indicating its advantage on watermark detection, too.

In addition, we observe larger performance drops for InternLM compared to Llama2 under the same watermarking method, indicating that impact of watermarking can vary among LLMs, highlighting the importance of model-specific evaluations.

In summary, while current watermarks are effective in detection, their generation quality still degrades significantly. Future work can explore new watermark designs to minimize such declines.

\subsection{Watermarking Strength Analysis}
\label{sec:level2_exp}

To analyze the influence of the watermarking strength on evaluating the detection performance and generation quality, we conduct experiments for the $4$ watermarking methods with $0.7$ watermarking strength at Table~\ref{tab:level345_strength2} to compare with the main results in Section~\ref{sec:Main_EXP} at $0.95$ watermarking strength. And we have the following observations: 

\noindent
(1) There exists a trade-off between the watermarking strength and generation quality. Models tend to exhibit larger drops in GM at $0.95$  strength compared to $0.7$. For example, the watermark with the worst generation score in Table~\ref{tab:level345_strength2} ($0.7$ strength) can rank the first in Table~\ref{tab:level345} ($0.95$ strength), which can't reflect the real difference between watermark algorithms. This highlights the importance of using a standardized strength for fair comparisons.

\noindent
(2)  At a lower strength of $0.7$ , average TP rates drop noticeably compared to $0.95$ strength across different tasks. We observe the largest TP rate drop (from $\sim$90\% to $\sim$0\%) in Category1 with short input and short answer. This suggests that our WaterBench is hard enough to add strong disturbance on watermarks for adjusting watermarking strengths.

\noindent
(3) V2 watermark maintains relatively stable detection and generation performance at both strengths, outperforming other methods. However, V2 watermark still makes Llama2's generation performance drops 31.2\%, indicating that further exploration is needed to minimize quality degradation.

\begin{figure}[t]
    \centering
    \includegraphics[width=\linewidth]{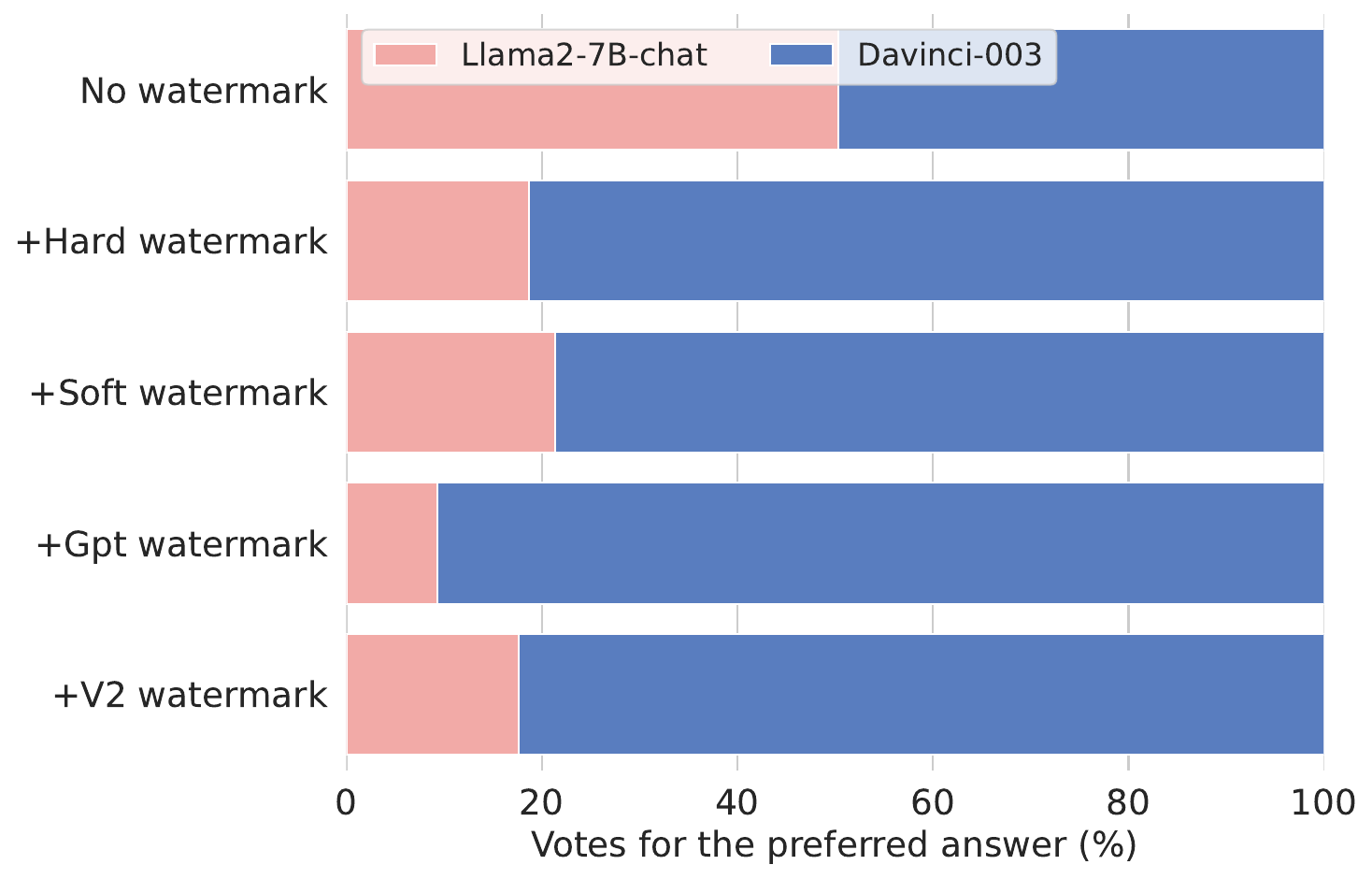}
    \caption{Average votes by three human annotators for the preferred answer between our watermarked LLM generation and text-davinci-003 baseline response.}
    \label{fig:votes}
\end{figure}

\subsection{Human Evaluation}
To prove the effectiveness of GPT4-Judge on task 5-1, we conduct a human evaluation that annotate the actual human preferences on model responses. We sample $100$ generation results respectively from the $5$ models at the watermarking strength level of $0.7$. Then we ask three human annotators to vote for their preferred response between the watermarked LLM and Davinci-003 baseline (See Appendix~\ref{appendix:human_detail} for more details). In total, we collect $1,500$ human feedback, yielding the following findings:

\noindent
(1) The results from the three human annotators align with the labeling from GPT4. Figure~\ref{fig:votes} exhibits the average voting results of three humans for the instruction-following task, where Llama2-7B-Chat without watermark achieves a 50.3\% win rate against the Davinci-003 baseline, which is consistent with the GPT4's simulated win rate of 54.7\%. Besides, the other watermarked LLMs also obtain the similar win rates to GPT4's predictions, further demonstrating the effectiveness of GPT4-Judge.

\begin{figure}[t]
    \centering
    \includegraphics[width=\linewidth]{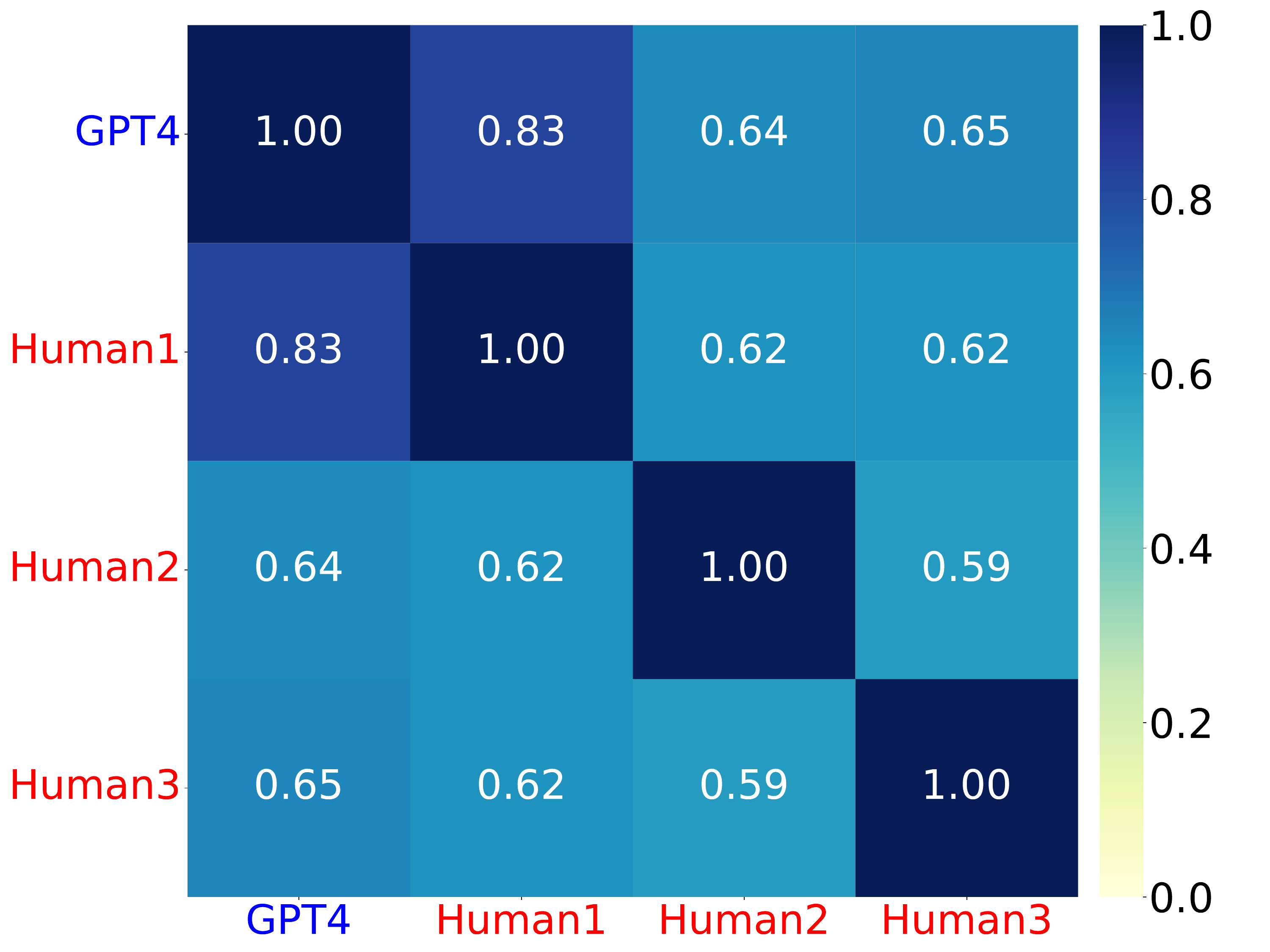}
    \caption{Cohen's kappa coefficient for inter-annotator agreement among GPT4 and human annotators.}
    \label{fig:IAA}
\end{figure}

\noindent
(2)  The inter-annotator agreement coefficients between GPT4 and three human annotators are varied, but all of them are above 0.6, indicating substantial agreement. As shown in Figure ~\ref{fig:IAA}, GPT4 has a 0.83 agreement with human1, which can be viewed as almost perfect agreement. While GPT4 only gets substantial agreement with human2 and human3. Additionally, the agreements among three human annotators are around 0.6, which means substantial agreement. So there also exists a variety of human annotators, which may lead to the different agreements with GPT4-Judge~\cite{dubois2023alpacafarm}.


\subsection{Correlation Analysis}

\begin{figure}[t]
    \centering
    \includegraphics[width=\linewidth]{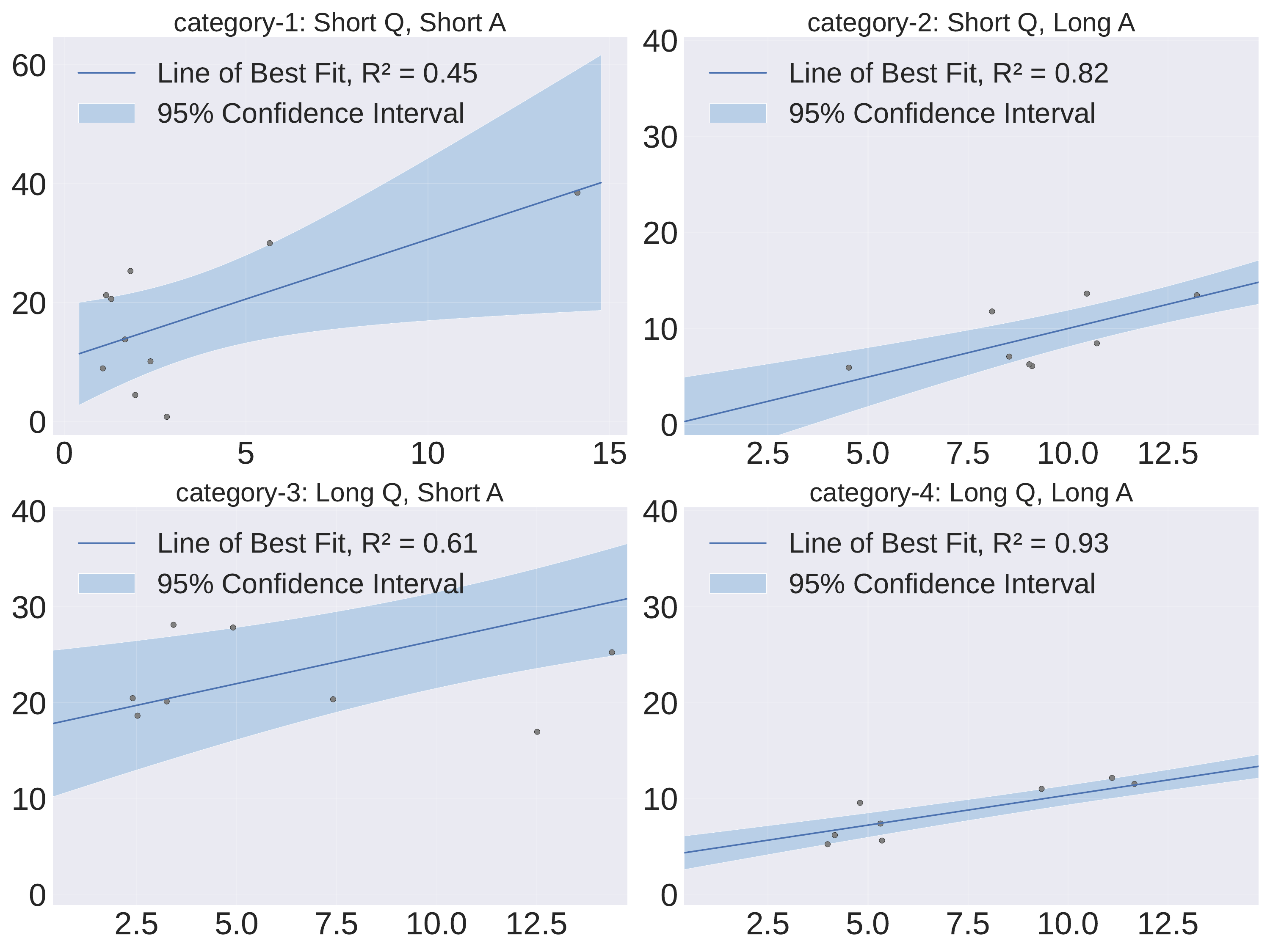}
    \caption{Scatter plots for each pair of tasks (e.g. 1-1 and 1-2), each point is an evaluated model's GM scores of two tasks in the same category.}
    \label{fig:length}
\end{figure}

To verify the diversity of our task selection, we analyze the inner task performance correlation for categories. 
As plotted in Figure~\ref{fig:length}, the generation performances of the watermarked LLMs on two sub-tasks of each category reveal a clear linear correlation, indicating the reliability of our task categorization. Notably, there is a more pronounced performance gap between tasks in the shorter answer categories (1 and 3). The convergence of task performances may reflect the model’s generalization capability on different tasks. Overall, WaterBench provides a comprehensive and challenging benchmark for evaluating LLM watermarks.

\section{Conclusion}
In this paper, we propose WaterBench, a new benchmark for evaluating large language model watermarks. We first introduce a benchmarking procedure that searches hyperparameters to ensure consistent watermarking strength across different methods, allowing for a fair comparison. Second, we construct a multi-task benchmark spanning nine typical NLP tasks with varying input/output lengths. Finally, we incorporate the GPT4-Judge metric to automatically evaluate the results. Experiments show that it sensitively reflects declines in instruction-following quality after watermarking.  We hope that our work will inspire and facilitate the future research on LLM watermarks.

\section*{Acknowledgement}
This work is supported by a grant from the Institute for Guo Qiang, Tsinghua University (2019GQB0003), Tsinghua University Initiative Scientific Research Program and Zhipu AI.

\section*{Limitations}
Although we have conducted extensive experiments, there are still some limitations for our work: (1) The detection candidate is only the reference answer in the benchmark, which is mainly written in the human expert style~\cite{ghosal2023towards}. However, all texts without the watermark can be considered as negative examples. (2) There is only one generation metric for each task. We will explore more metrics such as BertScore~\cite{zhang2019bertscore} and FactCC~\cite{kryscinski2019evaluating} to evaluate the  performance of LLMs in different aspects. (3)  A watermarking method may have different compositions for hyper-parameters to achieve the same watermarking strength, while in our experiments we only evaluate one composition with the minimum changes to  hyper-parameters. We encourage the future research to explore this composition.

\section*{Ethics Statement}

In this section, we will discuss the ethical consideration for our work.
\paragraph{Licenses.} For open-accessible datasets used in our work, we have checked their licenses. The KoLA~\cite{yu2023kola} dataset is shared under the GPLv3 license, Copen~\cite{peng2022copen} is shared under the MIT license, ELI5~\cite{fan-etal-2019-eli5} is shared under the BSD license, the LongBench~\cite{bai2023longbench} which includes task 3-1 to 4-2 is released under the MIT license, and the AlpacaFarm dataset~\cite{dubois2023alpacafarm} is shared under the Apache-2.0 license.
The Licenses for the large language models are also available.  Llama2-7B-chat~\cite{touvron2023llama} is released under the Meta License which needs to apply on their websites, and InternLM-7B-8k~\cite{2023internlm} is shared under the Apache-2.0 license.

\paragraph{Ethics Considerations for AI assistants} AI assistants like GPT4 are powerful, even our automatic evaluation process has adapted GPT4 as an evaluator, which complies with the AI ethical guidelines set by the European Union\footnote{\url{https://digital-strategy.ec.europa.eu/en/library/ethics-guidelines-trustworthy-ai}}. These guidelines place emphasis on various ethical aspects, including technical robustness, safety, privacy, transparency, and accountability. We make sure that the usage of AI systems in our research are aligned with these principles. They also highlight the importance of ensuring the safety  of AI systems and establishing accountability mechanisms for potential negative consequences. This encourages our work to evaluate LLM watermarks that may help policy makers conduct regulations for generative AI systems with detectable watermarks.

\bibliographystyle{acl_natbib}
\bibliography{custom}

\begin{thebibliography}{53}
\expandafter\ifx\csname natexlab\endcsname\relax\def\natexlab#1{#1}\fi

\bibitem[{Akhaee et~al.(2009)Akhaee, Sahraeian, Sankur, and Marvasti}]{akhaee2009robust}
Mohammad~Ali Akhaee, S~Mohammad~Ebrahim Sahraeian, Bulent Sankur, and Farokh Marvasti. 2009.
\newblock Robust scaling-based image watermarking using maximum-likelihood decoder with optimum strength factor.
\newblock \emph{IEEE Transactions on Multimedia}, 11(5):822--833.

\bibitem[{Alibrahim and Ludwig(2021)}]{alibrahim2021hyperparameter}
Hussain Alibrahim and Simone~A Ludwig. 2021.
\newblock Hyperparameter optimization: Comparing genetic algorithm against grid search and bayesian optimization.
\newblock In \emph{2021 IEEE Congress on Evolutionary Computation (CEC)}, pages 1551--1559. IEEE.

\bibitem[{Atallah et~al.(2001)Atallah, Raskin, Crogan, Hempelmann, Kerschbaum, Mohamed, and Naik}]{atallah2001natural}
Mikhail~J Atallah, Victor Raskin, Michael Crogan, Christian Hempelmann, Florian Kerschbaum, Dina Mohamed, and Sanket Naik. 2001.
\newblock Natural language watermarking: Design, analysis, and a proof-of-concept implementation.
\newblock In \emph{Information Hiding: 4th International Workshop, IH 2001 Pittsburgh, PA, USA, April 25--27, 2001 Proceedings 4}, pages 185--200. Springer.

\bibitem[{Bai et~al.(2023)Bai, Lv, Zhang, Lyu, Tang, Huang, Du, Liu, Zeng, Hou, Dong, Tang, and Li}]{bai2023longbench}
Yushi Bai, Xin Lv, Jiajie Zhang, Hongchang Lyu, Jiankai Tang, Zhidian Huang, Zhengxiao Du, Xiao Liu, Aohan Zeng, Lei Hou, Yuxiao Dong, Jie Tang, and Juanzi Li. 2023.
\newblock Longbench: A bilingual, multitask benchmark for long context understanding.
\newblock \emph{arXiv preprint arXiv:2308.14508}.

\bibitem[{Bubeck et~al.(2023)Bubeck, Chandrasekaran, Eldan, Gehrke, Horvitz, Kamar, Lee, Lee, Li, Lundberg et~al.}]{bubeck2023sparks}
S{\'e}bastien Bubeck, Varun Chandrasekaran, Ronen Eldan, Johannes Gehrke, Eric Horvitz, Ece Kamar, Peter Lee, Yin~Tat Lee, Yuanzhi Li, Scott Lundberg, et~al. 2023.
\newblock \href {https://arxiv.org/abs/2303.12712} {Sparks of artificial general intelligence: Early experiments with gpt-4}.
\newblock \emph{arXiv preprint arXiv:2303.12712}.

\bibitem[{Cai et~al.(2023)Cai, Haslett, Duan, Wang, and Pickering}]{cai2023does}
Zhenguang~G Cai, David~A Haslett, Xufeng Duan, Shuqi Wang, and Martin~J Pickering. 2023.
\newblock Does chatgpt resemble humans in language use?
\newblock \emph{arXiv preprint arXiv:2303.08014}.

\bibitem[{Chen et~al.(2021)Chen, Tworek, Jun, Yuan, Pinto, Kaplan, Edwards, Burda, Joseph, Brockman et~al.}]{chen2021evaluating}
Mark Chen, Jerry Tworek, Heewoo Jun, Qiming Yuan, Henrique Ponde de~Oliveira Pinto, Jared Kaplan, Harri Edwards, Yuri Burda, Nicholas Joseph, Greg Brockman, et~al. 2021.
\newblock Evaluating large language models trained on code.
\newblock \emph{arXiv preprint arXiv:2107.03374}.

\bibitem[{Chia et~al.(2023)Chia, Hong, Bing, and Poria}]{chia2023instructeval}
Yew~Ken Chia, Pengfei Hong, Lidong Bing, and Soujanya Poria. 2023.
\newblock \href {https://arxiv.org/abs/2306.04757} {{INSTRUCTEVAL}: Towards holistic evaluation of instruction-tuned large language models}.
\newblock \emph{arXiv preprint arXiv:2306.04757}.

\bibitem[{Dubois et~al.(2023)Dubois, Li, Taori, Zhang, Gulrajani, Ba, Guestrin, Liang, and Hashimoto}]{dubois2023alpacafarm}
Yann Dubois, Xuechen Li, Rohan Taori, Tianyi Zhang, Ishaan Gulrajani, Jimmy Ba, Carlos Guestrin, Percy Liang, and Tatsunori~B Hashimoto. 2023.
\newblock Alpacafarm: A simulation framework for methods that learn from human feedback.
\newblock \emph{arXiv preprint arXiv:2305.14387}.

\bibitem[{Fabbri et~al.(2019)Fabbri, Li, She, Li, and Radev}]{fabbri2019multi}
Alexander~Richard Fabbri, Irene Li, Tianwei She, Suyi Li, and Dragomir Radev. 2019.
\newblock Multi-news: A large-scale multi-document summarization dataset and abstractive hierarchical model.
\newblock In \emph{Proceedings of the 57th Annual Meeting of the Association for Computational Linguistics}, pages 1074--1084.

\bibitem[{Fan et~al.(2019)Fan, Jernite, Perez, Grangier, Weston, and Auli}]{fan-etal-2019-eli5}
Angela Fan, Yacine Jernite, Ethan Perez, David Grangier, Jason Weston, and Michael Auli. 2019.
\newblock \href {https://doi.org/10.18653/v1/P19-1346} {{ELI}5: Long form question answering}.
\newblock In \emph{Proceedings of ACL}, pages 3558--3567.

\bibitem[{Fernandez et~al.(2023)Fernandez, Chaffin, Tit, Chappelier, and Furon}]{fernandez2023three}
Pierre Fernandez, Antoine Chaffin, Karim Tit, Vivien Chappelier, and Teddy Furon. 2023.
\newblock Three bricks to consolidate watermarks for large language models.
\newblock \emph{arXiv preprint arXiv:2308.00113}.

\bibitem[{Fu et~al.(2023)Fu, Xiong, and Dong}]{fu2023watermarking}
Yu~Fu, Deyi Xiong, and Yue Dong. 2023.
\newblock Watermarking conditional text generation for ai detection: Unveiling challenges and a semantic-aware watermark remedy.
\newblock \emph{arXiv preprint arXiv:2307.13808}.

\bibitem[{Ghosal et~al.(2023)Ghosal, Chakraborty, Geiping, Huang, Manocha, and Bedi}]{ghosal2023towards}
Soumya~Suvra Ghosal, Souradip Chakraborty, Jonas Geiping, Furong Huang, Dinesh Manocha, and Amrit~Singh Bedi. 2023.
\newblock Towards possibilities \& impossibilities of ai-generated text detection: A survey.
\newblock \emph{arXiv preprint arXiv:2310.15264}.

\bibitem[{Guo et~al.(2023)Guo, Zhang, Wang, Jiang, Nie, Ding, Yue, and Wu}]{guo2023close}
Biyang Guo, Xin Zhang, Ziyuan Wang, Minqi Jiang, Jinran Nie, Yuxuan Ding, Jianwei Yue, and Yupeng Wu. 2023.
\newblock \href {https://arxiv.org/abs/2301.07597} {How close is chatgpt to human experts? comparison corpus, evaluation, and detection}.
\newblock \emph{arXiv preprint arXiv:2301.07597}.

\bibitem[{Hou et~al.(2024)Hou, Zhang, Wang, Khashabi, and He}]{hou2024ksemstamp}
Abe~Bohan Hou, Jingyu Zhang, Yichen Wang, Daniel Khashabi, and Tianxing He. 2024.
\newblock \href {http://arxiv.org/abs/2402.11399} {k-semstamp: A clustering-based semantic watermark for detection of machine-generated text}.

\bibitem[{Hu et~al.(2023)Hu, Chen, Wu, Wu, Zhang, and Huang}]{hu2023unbiased}
Zhengmian Hu, Lichang Chen, Xidong Wu, Yihan Wu, Hongyang Zhang, and Heng Huang. 2023.
\newblock Unbiased watermark for large language models.
\newblock \emph{arXiv preprint arXiv:2310.10669}.

\bibitem[{Kirchenbauer et~al.(2023{\natexlab{a}})Kirchenbauer, Geiping, Wen, Katz, Miers, and Goldstein}]{kirchenbauer2023watermark}
John Kirchenbauer, Jonas Geiping, Yuxin Wen, Jonathan Katz, Ian Miers, and Tom Goldstein. 2023{\natexlab{a}}.
\newblock A watermark for large language models.
\newblock \emph{arXiv preprint arXiv:2301.10226}.

\bibitem[{Kirchenbauer et~al.(2023{\natexlab{b}})Kirchenbauer, Geiping, Wen, Shu, Saifullah, Kong, Fernando, Saha, Goldblum, and Goldstein}]{kirchenbauer2023reliability}
John Kirchenbauer, Jonas Geiping, Yuxin Wen, Manli Shu, Khalid Saifullah, Kezhi Kong, Kasun Fernando, Aniruddha Saha, Micah Goldblum, and Tom Goldstein. 2023{\natexlab{b}}.
\newblock On the reliability of watermarks for large language models.
\newblock \emph{arXiv preprint arXiv:2306.04634}.

\bibitem[{Krishna et~al.(2023)Krishna, Song, Karpinska, Wieting, and Iyyer}]{krishna2023paraphrasing}
Kalpesh Krishna, Yixiao Song, Marzena Karpinska, John Wieting, and Mohit Iyyer. 2023.
\newblock \href {https://arxiv.org/abs/2303.13408} {Paraphrasing evades detectors of ai-generated text, but retrieval is an effective defense}.
\newblock \emph{arXiv preprint arXiv:2303.13408}.

\bibitem[{Kry{\'s}ci{\'n}ski et~al.(2019)Kry{\'s}ci{\'n}ski, McCann, Xiong, and Socher}]{kryscinski2019evaluating}
Wojciech Kry{\'s}ci{\'n}ski, Bryan McCann, Caiming Xiong, and Richard Socher. 2019.
\newblock Evaluating the factual consistency of abstractive text summarization.
\newblock \emph{arXiv preprint arXiv:1910.12840}.

\bibitem[{Kuditipudi et~al.(2023)Kuditipudi, Thickstun, Hashimoto, and Liang}]{kuditipudi2023robust}
Rohith Kuditipudi, John Thickstun, Tatsunori Hashimoto, and Percy Liang. 2023.
\newblock Robust distortion-free watermarks for language models.
\newblock \emph{arXiv preprint arXiv:2307.15593}.

\bibitem[{Li et~al.(2023{\natexlab{a}})Li, Guo, Fan, Xu, and Song}]{li2023multi}
Haoran Li, Dadi Guo, Wei Fan, Mingshi Xu, and Yangqiu Song. 2023{\natexlab{a}}.
\newblock Multi-step jailbreaking privacy attacks on chatgpt.
\newblock \emph{arXiv preprint arXiv:2304.05197}.

\bibitem[{Li et~al.(2023{\natexlab{b}})Li, Wang, Shi, and Hsieh}]{li2023improving}
Yuhang Li, Yihan Wang, Zhouxing Shi, and Cho-Jui Hsieh. 2023{\natexlab{b}}.
\newblock \href {http://arxiv.org/abs/2311.09668} {Improving the generation quality of watermarked large language models via word importance scoring}.

\bibitem[{Liu et~al.(2023)Liu, Pan, Hu, Meng, and Wen}]{liu2023semantic}
Aiwei Liu, Leyi Pan, Xuming Hu, Shiao Meng, and Lijie Wen. 2023.
\newblock A semantic invariant robust watermark for large language models.
\newblock \emph{arXiv preprint arXiv:2310.06356}.

\bibitem[{Liu et~al.(2024)Liu, Pan, Lu, Li, Hu, Zhang, Wen, King, Xiong, and Yu}]{liu2024survey}
Aiwei Liu, Leyi Pan, Yijian Lu, Jingjing Li, Xuming Hu, Xi~Zhang, Lijie Wen, Irwin King, Hui Xiong, and Philip~S. Yu. 2024.
\newblock \href {http://arxiv.org/abs/2312.07913} {A survey of text watermarking in the era of large language models}.

\bibitem[{Lu et~al.(2024)Lu, Liu, Yu, Li, and King}]{lu2024entropybased}
Yijian Lu, Aiwei Liu, Dianzhi Yu, Jingjing Li, and Irwin King. 2024.
\newblock \href {http://arxiv.org/abs/2403.13485} {An entropy-based text watermarking detection method}.

\bibitem[{Maia et~al.(2018)Maia, Handschuh, Freitas, Davis, McDermott, Zarrouk, and Balahur}]{maia201818}
Macedo Maia, Siegfried Handschuh, Andr{\'e} Freitas, Brian Davis, Ross McDermott, Manel Zarrouk, and Alexandra Balahur. 2018.
\newblock Www'18 open challenge: financial opinion mining and question answering.
\newblock In \emph{Companion proceedings of the the web conference 2018}, pages 1941--1942.

\bibitem[{Mei et~al.(2002)Mei, Li, Dang, and Wang}]{mei2002decision}
Shi-chun Mei, Ren-hou Li, Hong-mei Dang, and Yun-kuan Wang. 2002.
\newblock Decision of image watermarking strength based on artificial neural-networks.
\newblock In \emph{Proceedings of the 9th International Conference on Neural Information Processing, 2002. ICONIP'02.}, volume~5, pages 2430--2434. IEEE.

\bibitem[{Mitchell et~al.(2023)Mitchell, Lee, Khazatsky, Manning, and Finn}]{mitchell2023detectgpt}
Eric Mitchell, Yoonho Lee, Alexander Khazatsky, Christopher~D Manning, and Chelsea Finn. 2023.
\newblock \href {https://arxiv.org/abs/2301.11305} {Detectgpt: Zero-shot machine-generated text detection using probability curvature}.
\newblock \emph{arXiv preprint arXiv:2301.11305}.

\bibitem[{Munyer and Zhong(2023)}]{munyer2023deeptextmark}
Travis Munyer and Xin Zhong. 2023.
\newblock Deeptextmark: Deep learning based text watermarking for detection of large language model generated text.
\newblock \emph{arXiv preprint arXiv:2305.05773}.

\bibitem[{OpenAI(2023)}]{openai2023gpt4}
OpenAI. 2023.
\newblock Gpt-4 technical report,.
\newblock \emph{OpenAI}.

\bibitem[{Pan et~al.(2024)Pan, Liu, He, Gao, Zhao, Lu, Zhou, Liu, Hu, Wen, and King}]{pan2024markllm}
Leyi Pan, Aiwei Liu, Zhiwei He, Zitian Gao, Xuandong Zhao, Yijian Lu, Binglin Zhou, Shuliang Liu, Xuming Hu, Lijie Wen, and Irwin King. 2024.
\newblock \href {http://arxiv.org/abs/2405.10051} {Markllm: An open-source toolkit for llm watermarking}.

\bibitem[{Peng et~al.(2022)Peng, Wang, Hu, Jin, Hou, Li, Liu, and Liu}]{peng2022copen}
Hao Peng, Xiaozhi Wang, Shengding Hu, Hailong Jin, Lei Hou, Juanzi Li, Zhiyuan Liu, and Qun Liu. 2022.
\newblock Copen: Probing conceptual knowledge in pre-trained language models.
\newblock \emph{arXiv preprint arXiv:2211.04079}.

\bibitem[{Raffel et~al.(2020)Raffel, Shazeer, Roberts, Lee, Narang, Matena, Zhou, Li, and Liu}]{raffel2020exploring}
Colin Raffel, Noam Shazeer, Adam Roberts, Katherine Lee, Sharan Narang, Michael Matena, Yanqi Zhou, Wei Li, and Peter~J Liu. 2020.
\newblock Exploring the limits of transfer learning with a unified text-to-text transformer.
\newblock \emph{The Journal of Machine Learning Research}, 21(1):5485--5551.

\bibitem[{Sadasivan et~al.(2023)Sadasivan, Kumar, Balasubramanian, Wang, and Feizi}]{sadasivan2023can}
Vinu~Sankar Sadasivan, Aounon Kumar, Sriram Balasubramanian, Wenxiao Wang, and Soheil Feizi. 2023.
\newblock Can ai-generated text be reliably detected?
\newblock \emph{arXiv preprint arXiv:2303.11156}.

\bibitem[{Sato et~al.(2023)Sato, Takezawa, Bao, Niwa, and Yamada}]{sato2023embarrassingly}
Ryoma Sato, Yuki Takezawa, Han Bao, Kenta Niwa, and Makoto Yamada. 2023.
\newblock Embarrassingly simple text watermarks.
\newblock \emph{arXiv preprint arXiv:2310.08920}.

\bibitem[{Takezawa et~al.(2023)Takezawa, Sato, Bao, Niwa, and Yamada}]{takezawa2023necessary}
Yuki Takezawa, Ryoma Sato, Han Bao, Kenta Niwa, and Makoto Yamada. 2023.
\newblock Necessary and sufficient watermark for large language models.
\newblock \emph{arXiv preprint arXiv:2310.00833}.

\bibitem[{Tang et~al.(2023)Tang, Chuang, and Hu}]{tang2023science}
Ruixiang Tang, Yu-Neng Chuang, and Xia Hu. 2023.
\newblock \href {https://arxiv.org/abs/2303.07205} {The science of detecting llm-generated texts}.
\newblock \emph{arXiv preprint arXiv:2303.07205}.

\bibitem[{Team(2023)}]{2023internlm}
InternLM Team. 2023.
\newblock Internlm: A multilingual language model with progressively enhanced capabilities.
\newblock \url{https://github.com/InternLM/InternLM}.

\bibitem[{Topkara et~al.(2005)Topkara, Taskiran, and Delp~III}]{topkara2005natural}
Mercan Topkara, Cuneyt~M Taskiran, and Edward~J Delp~III. 2005.
\newblock Natural language watermarking.
\newblock In \emph{Security, Steganography, and Watermarking of Multimedia Contents VII}, volume 5681, pages 441--452. SPIE.

\bibitem[{Touvron et~al.(2023)Touvron, Martin, Stone, Albert, Almahairi, Babaei, Bashlykov, Batra, Bhargava, Bhosale et~al.}]{touvron2023llama}
Hugo Touvron, Louis Martin, Kevin Stone, Peter Albert, Amjad Almahairi, Yasmine Babaei, Nikolay Bashlykov, Soumya Batra, Prajjwal Bhargava, Shruti Bhosale, et~al. 2023.
\newblock Llama 2: Open foundation and fine-tuned chat models.
\newblock \emph{arXiv preprint arXiv:2307.09288}.

\bibitem[{Tu et~al.(2023)Tu, Li, Yu, Wang, Hou, and Li}]{tu2023chatlog}
Shangqing Tu, Chunyang Li, Jifan Yu, Xiaozhi Wang, Lei Hou, and Juanzi Li. 2023.
\newblock Chatlog: Recording and analyzing chatgpt across time.
\newblock \emph{arXiv preprint arXiv:2304.14106}.

\bibitem[{Wang et~al.(2023{\natexlab{a}})Wang, Li, Chen, Zhu, Lin, Cao, Liu, Liu, and Sui}]{wang2023large}
Peiyi Wang, Lei Li, Liang Chen, Dawei Zhu, Binghuai Lin, Yunbo Cao, Qi~Liu, Tianyu Liu, and Zhifang Sui. 2023{\natexlab{a}}.
\newblock Large language models are not fair evaluators.
\newblock \emph{arXiv preprint arXiv:2305.17926}.

\bibitem[{Wang et~al.(2023{\natexlab{b}})Wang, Cheng, Cui, and Yu}]{wang2023implementing}
Zecong Wang, Jiaxi Cheng, Chen Cui, and Chenhao Yu. 2023{\natexlab{b}}.
\newblock Implementing bert and fine-tuned roberta to detect ai generated news by chatgpt.
\newblock \emph{arXiv preprint arXiv:2306.07401}.

\bibitem[{Yang et~al.(2023)Yang, Chen, Zhang, Liu, Qi, Zhang, Fang, and Yu}]{yang2023watermarking}
Xi~Yang, Kejiang Chen, Weiming Zhang, Chang Liu, Yuang Qi, Jie Zhang, Han Fang, and Nenghai Yu. 2023.
\newblock Watermarking text generated by black-box language models.
\newblock \emph{arXiv preprint arXiv:2305.08883}.

\bibitem[{Yang et~al.(2018)Yang, Qi, Zhang, Bengio, Cohen, Salakhutdinov, and Manning}]{yang2018hotpotqa}
Zhilin Yang, Peng Qi, Saizheng Zhang, Yoshua Bengio, William Cohen, Ruslan Salakhutdinov, and Christopher~D Manning. 2018.
\newblock Hotpotqa: A dataset for diverse, explainable multi-hop question answering.
\newblock In \emph{Proceedings of the 2018 Conference on Empirical Methods in Natural Language Processing}, pages 2369--2380.

\bibitem[{Yoo et~al.(2023)Yoo, Ahn, Jang, and Kwak}]{yoo2023robust}
KiYoon Yoo, Wonhyuk Ahn, Jiho Jang, and Nojun Kwak. 2023.
\newblock Robust natural language watermarking through invariant features.
\newblock \emph{arXiv preprint arXiv:2305.01904}.

\bibitem[{Yu et~al.(2023)Yu, Wang, Tu, Cao, Zhang-Li, Lv, Peng, Yao, Zhang, Li et~al.}]{yu2023kola}
Jifan Yu, Xiaozhi Wang, Shangqing Tu, Shulin Cao, Daniel Zhang-Li, Xin Lv, Hao Peng, Zijun Yao, Xiaohan Zhang, Hanming Li, et~al. 2023.
\newblock Kola: Carefully benchmarking world knowledge of large language models.
\newblock \emph{arXiv preprint arXiv:2306.09296}.

\bibitem[{Zhang et~al.(2019)Zhang, Kishore, Wu, Weinberger, and Artzi}]{zhang2019bertscore}
Tianyi Zhang, Varsha Kishore, Felix Wu, Kilian~Q Weinberger, and Yoav Artzi. 2019.
\newblock Bertscore: Evaluating text generation with bert.
\newblock \emph{arXiv preprint arXiv:1904.09675}.

\bibitem[{Zhao et~al.(2023)Zhao, Ananth, Li, and Wang}]{zhao2023provable}
Xuandong Zhao, Prabhanjan Ananth, Lei Li, and Yu-Xiang Wang. 2023.
\newblock Provable robust watermarking for ai-generated text.
\newblock \emph{arXiv preprint arXiv:2306.17439}.

\bibitem[{Zheng et~al.(2023)Zheng, Chiang, Sheng, Zhuang, Wu, Zhuang, Lin, Li, Li, Xing et~al.}]{zheng2023judging}
Lianmin Zheng, Wei-Lin Chiang, Ying Sheng, Siyuan Zhuang, Zhanghao Wu, Yonghao Zhuang, Zi~Lin, Zhuohan Li, Dacheng Li, Eric Xing, et~al. 2023.
\newblock Judging llm-as-a-judge with mt-bench and chatbot arena.
\newblock \emph{arXiv preprint arXiv:2306.05685}.

\bibitem[{Zhong et~al.(2021)Zhong, Yin, Yu, Zaidi, Mutuma, Jha, Hassan, Celikyilmaz, Liu, Qiu et~al.}]{zhong2021qmsum}
Ming Zhong, Da~Yin, Tao Yu, Ahmad Zaidi, Mutethia Mutuma, Rahul Jha, Ahmed Hassan, Asli Celikyilmaz, Yang Liu, Xipeng Qiu, et~al. 2021.
\newblock Qmsum: A new benchmark for query-based multi-domain meeting summarization.
\newblock In \emph{Proceedings of the 2021 Conference of the North American Chapter of the Association for Computational Linguistics: Human Language Technologies}, pages 5905--5921.

\end{thebibliography}

\appendix

\section{Implementation Details}
\label{sec:implementation}

\subsection{Deployment Details}
In our evaluating and detecting experiments, we utilize the widely-used \textit{Pytorch} and \textit{transformers} library to load all the models. All the experiments are conducted on Ubuntu 20.04.4 server equipped with 112 Intel Xeon(R) Platinum 8336C CPU cores, and graphic cards that contained 8 NVIDIA
A100 SXM 80GB GPUs. Besides, the CUDA version is 11.4, the Python version is 3.10.11, the
PyTorch version is 2.0.1 and the transformers version is 4.31.0. We integrate the code from LM-Watermark\footnote{ \url{https://github.com/jwkirchenbauer/lm-watermarking}}, V2 Watermark\footnote{ \url{https://github.com/jwkirchenbauer/lm-watermarking/tree/main/watermark_reliability_release}} and GPT Watermark\footnote{\url{https://github.com/XuandongZhao/GPTWatermark}} to implement a unified watermarking experiment tool, where different kinds of watermark can be evaluated fairly. The code of our all-in-one tool is provided in the supplement files.


\subsection{Hyper-parameters Search Details}
\label{sec:search_ruc_detail}
In order to obtain experimental groups with the same watermark strength, there are three hyper-parameters that need to be obtained through search. The first is the vocabulary partition parameter $\gamma$, which represents the proportion of the green list vocabulary within the model's total vocabulary. The second is the bias constant $\delta$ for the logit, representing the hardness of the red list vocabulary. And the last one is the threshold used in the Z-test. Under the same threshold conditions, according to the calculation method of the $z$-score, it can be observed in Figure~\ref{fig:hyper_res} that when $\gamma$ increases, the average $z$-score will decrease, resulting in a weaker watermark strength. And the increase in $\delta$ implies a stronger hardness of the watermark, which in turn results in a stronger watermark strength. Therefore, we first set the same threshold and adjust these two hyper-parameters, $\gamma$ and $\delta$, based on their relationship with watermark strength to find different watermark groups with the same strength category. Then, we make slight adjustments to the threshold for these watermark groups to ensure they achieve the same strength level with greater precision. Using this approach, we obtained the appropriate hyper-parameters and recorded them in Table 6. Note that the default $z$-score threshold is $4$, which is  a commonly used value in prior works~\cite{kirchenbauer2023watermark,kirchenbauer2023reliability}.


\begin{figure} [ht!]
	\centering
	\subfloat[\label{2a}]{
		\includegraphics[scale=0.5]{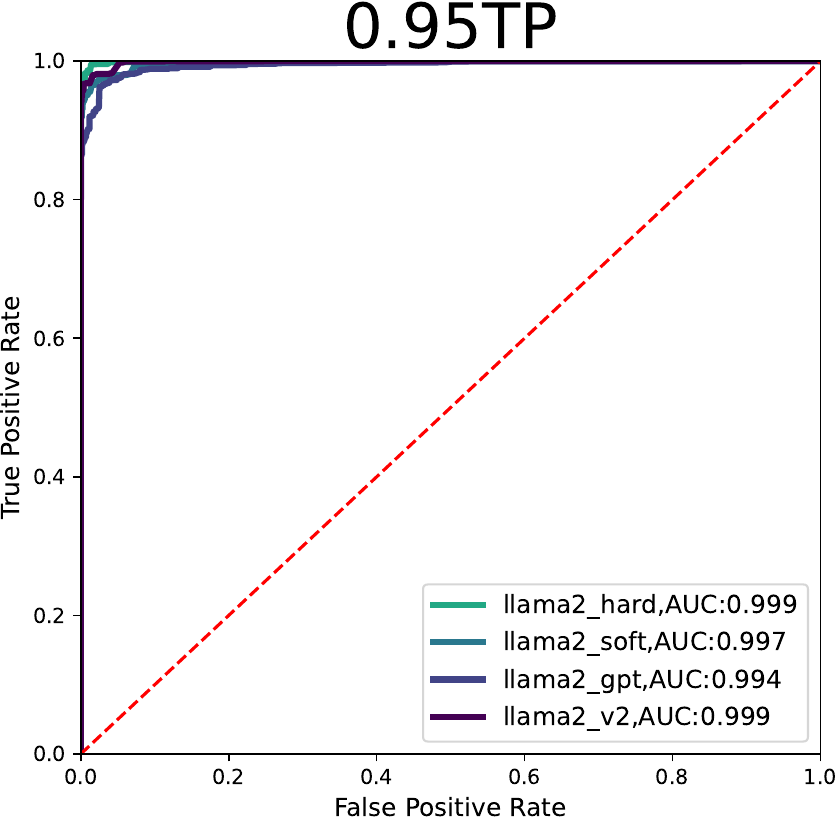}}
	\\
	\subfloat[\label{2b}]{
		\includegraphics[scale=0.5]{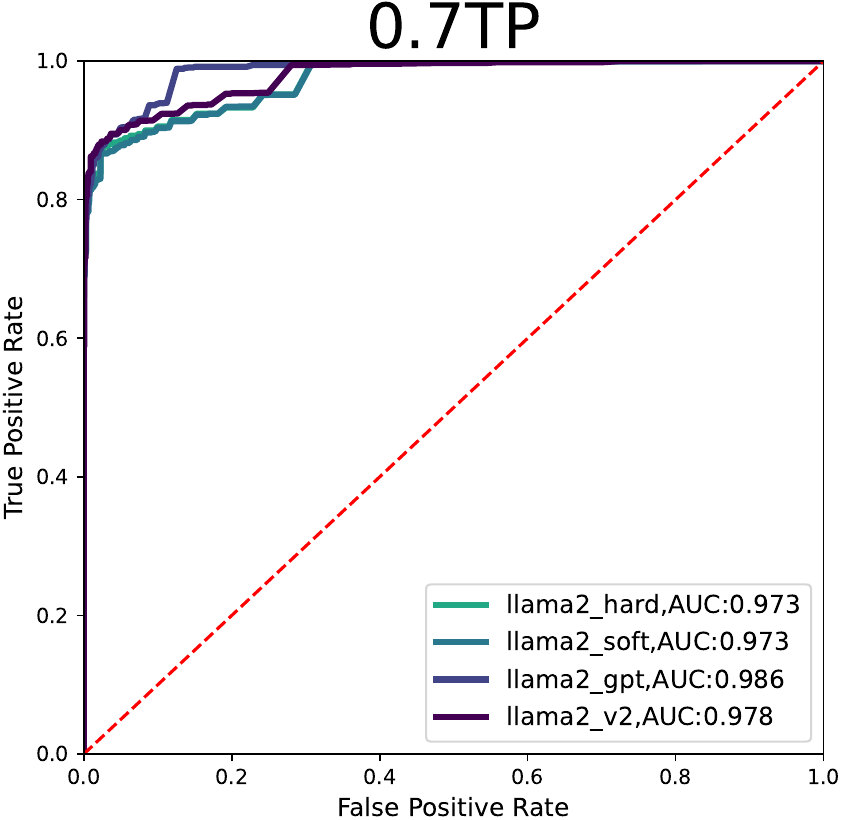} }
	\caption{ROC curves for two watermark strengths.}
	\label{roc_auc} 
\end{figure}

To prove the effectiveness of our hyper-parameter searching process, we draw the ROC curve in Figure~\ref{roc_auc} by adjusting the $z$-score threshold. First, all of the $4$ kinds of watermarks on the same watermark level obtain similar AUC scores, where even all AUC socres are 0.99 at the 0.95 watermark strength. As there are only small difference among watermarks on the ROC curve, it is reasonable for us to set the same initial $z$-score threshold for all models. Second, when at the perfect point that when False Positive Rate is $0$, the True Positive Rate is often not at the target level for some watermarks. That's why we need to adjust the $z$-score threshold to reach the watermark strength in the end. Finally, as we first adjust the $\gamma$ and $\delta$ to get different watermark strengths, we can observe the difference of 0.7 and 0.95 strengths caused by these hyper-parameters on the sub-figures in Figure~\ref{roc_auc}. If we fix the $\gamma$ and $\delta$, and only adjust the $z$-score threshold, then we may not get the ideal TPR that can be reached by adjusting $\gamma$ and $\delta$.

\begin{table}[ht]

    \centering
    \resizebox{0.94\linewidth}{!}{
    \begin{tabular}{c|c|c}
        \toprule
\textbf{Module}   &   \textbf{Parameter} & \textbf{Value}   \\
                \midrule 
         
  \multirow{2}{*}{ \shortstack{Llama2 + \\hard watermark} }  
  & $\gamma$   & 0.25   \\
  & 0.95TP z-score threshold & 4.3   \\
         \midrule

    \multirow{3}{*}{ \shortstack{Llama2 + \\soft watermark} }  
  & $\gamma$   & 0.1   \\
  & $\delta$  & 10   \\
  & 0.95TP z-score threshold & 4.0   \\
         \midrule

    \multirow{3}{*}{ \shortstack{Llama2 + \\gpt watermark} }  
  & $\gamma$   & 0.1   \\
  & $\delta$  & 10   \\
  & 0.95TP z-score threshold & 4.2   \\
         \midrule

    \multirow{3}{*}{ \shortstack{Llama2 + \\v2 watermark} }  
  & $\gamma$   & 0.25   \\
  & $\delta$  & 15   \\
  & 0.95TP z-score threshold & 4.1   \\
         \midrule

    \multirow{2}{*}{ \shortstack{Llama2 + \\hard watermark} }  
  & $\gamma$   & 0.75   \\
  & 0.7TP z-score threshold & 4.2   \\
         \midrule

    \multirow{3}{*}{ \shortstack{Llama2 + \\soft watermark} }  
  & $\gamma$   & 0.75   \\
  & $\delta$  & 15   \\
  & 0.7TP z-score threshold & 4.2   \\
         \midrule

    \multirow{3}{*}{ \shortstack{Llama2 + \\gpt watermark} }  
  & $\gamma$   & 0.65   \\
  & $\delta$  & 12.5   \\
  & 0.7TP z-score threshold & 4.0   \\
         \midrule

    \multirow{3}{*}{ \shortstack{Llama2 + \\v2 watermark} }  
  & $\gamma$   & 0.75   \\
  & $\delta$  & 15   \\
  & 0.7TP z-score threshold & 3.8   \\
         \midrule

    \multirow{2}{*}{ \shortstack{Internlm + \\hard watermark} }  
  & $\gamma$   & 0.15   \\
  & 0.95TP z-score threshold & 3.5   \\
         \midrule

    \multirow{3}{*}{ \shortstack{Internlm + \\soft watermark} }  
  & $\gamma$   & 0.1   \\
  & $\delta$  & 10   \\
  & 0.95TP z-score threshold & 3.2   \\
         \midrule

    \multirow{3}{*}{ \shortstack{Internlm + \\gpt watermark} }  
  & $\gamma$   & 0.25   \\
  & $\delta$  & 15   \\
  & 0.95TP z-score threshold & 4.1   \\
         \midrule

    \multirow{3}{*}{ \shortstack{Internlm + \\v2 watermark} }  
  & $\gamma$   & 0.1   \\
  & $\delta$  & 10   \\
  & 0.95TP z-score threshold & 4.0   \\
         \bottomrule

    \end{tabular}
        }
              \caption{Hyper-parameters for each model.  }
    \label{tab:hyper_parameters}
\end{table}

It is noted that there may exist many points that satisfy the fixed TPR, or quite close to the value, to select  hyper-parameters, we adopt the following approach:

First, we use a grid search approach to find the proper points. As shown in Table~\ref{tab:select1} and Table~\ref{tab:select2}, there may be many points proximate to a TPR of 0.95, albeit not precisely equal to it. We choose the points that exhibit the least deviation, like TPR=0.949 to report.

Subsequently, we examine two soft watermark results that approximated a TPR of 0.95, as derived from the hyperparameter search. For the overall scores, both two watermarks are around the level of TPR=0.95. Their TNR scores are even the same as 0.995 while their GM scores are a little different, one is 10.7 and the other is 11. Despite the disparity in their GM scores on C1 and C2, the scores on C3 and C4 are akin. Therefore, we generally ensure minimal differences exist between points around a TPR of 0.95. We choose to report the TPR=0.949 one over TPR=0.967.

We acknowledge that a more comprehensive comparison would involve analyzing the Pareto frontier of "TPR @ fixed TNR" versus "GM". This would provide a more accurate assessment of the trade-offs between these metrics across the different methods. However, given the need for extensibility in adding new watermarks and the computational cost of GPU resources, our current evaluation framework may not be able to accommodate the Pareto frontier analysis. 
  \begin{table*}[t]
  \vspace*{-2.0em}
  \centering 
  \resizebox{\textwidth}{!}{
  \begin{tabular}{l|cccc|cccc|cccc}
  \toprule
  \multirow{3}{*}{\textbf{Model}} & \multicolumn{4}{c|}{\textbf{C1: (Short Q, Short A)}} & \multicolumn{4}{c|}{\textbf{C2: (Short Q, Long A)}} &  \multicolumn{4}{c}{\textbf{C3: (Long Q, Short A)}}  \\
  & \multicolumn{4}{c|}{\textit{Factual Knowledge}} & \multicolumn{4}{c|}{\textit{Long-form QA} } & \multicolumn{4}{c}{\textit{Reasoning \& Coding}} \\
 
  & \textbf{TP} & \textbf{TN} & \textbf{GM} & \textbf{Drop} & \textbf{TP} & \textbf{TN} & \textbf{GM} & \textbf{Drop} & \textbf{TP} & \textbf{TN} & \textbf{GM} & \textbf{Drop} \\
  \midrule
  \rowcolor{mygray} Llama2-7b-chat   & -- & -- & 17.8 & -- & -- & -- & 21.3 & -- & -- & -- & 37.5 & -- \\
 + soft watermark $\gamma$=0.1, $\delta$=10  & 90.2 & 99.2 & 7.7 & $\downarrow$ 56.6\% & 100.0 & 100.0 & 9.9 & $\downarrow$ 53.3\% & 80.2 & 100.0 & 19.8 & $\downarrow$ 47.2\% \\
+ soft watermark $\gamma$=0.25, $\delta$=15 & 95.5 & 100.0 & 4.7 & $\downarrow$ 73.4\% & 100.0 & 99.8 & 13.0 & $\downarrow$ 38.7\% & 84.8 & 100.0 & 19.3 & $\downarrow$ 48.6\%\\
  
  \bottomrule
  \end{tabular}
  }
  \caption{True Positive Rate (TP), True Negative Rate (TN), Generation Metric (GM) and Generation Quality Drop (Drop) for category 4, 5 and all tasks at the watermarking strength level of 0.95 with $z$-score threshold of 4.}
  \label{tab:select1}
  \end{table*}

  \begin{table*}[t]
  \vspace*{-2.0em}
  \centering 
  \resizebox{\textwidth}{!}{
  \begin{tabular}{l|cccc|cccc|cccc}
  \toprule
  
  \multirow{3}{*}{\textbf{Model}} & \multicolumn{4}{c|}{\textbf{C4: (Long Q, Long A)}} & \multicolumn{4}{c|}{\textbf{C5: Open-Ended}} &  \multicolumn{4}{c}{\textbf{Overall: (12345)}}  \\
  & \multicolumn{4}{c|}{\textit{Summarization}} & \multicolumn{4}{c|}{\textit{Instruction Following} } & \multicolumn{4}{c}{\textit{Detection
  \& Generation }} \\
  & \textbf{TP} & \textbf{TN} & \textbf{GM} & \textbf{Drop} & \textbf{TP} & \textbf{TN} & \textbf{GM} & \textbf{Drop} & \textbf{TP} & \textbf{TN} & \textbf{GM} & \textbf{Drop} \\
  \midrule
  \rowcolor{mygray} Llama2-7b-chat   & -- & -- & 23.3 & -- & -- & -- & 54.7 & -- & -- & -- & 28.3 & -- \\
+ soft watermark $\gamma$=0.1, $\delta$=10  & 100.0 & 100.0 & 10.2 & $\downarrow$ 56.3\% & 99.4 & 98.9 & 0.6 & $\downarrow$ 98.9\% & 94.9 & 99.5 & 10.7 & $\downarrow$ 62.3\% \\
+ soft watermark $\gamma$=0.25, $\delta$=15 & 100.0 & 100.0 & 12.2 & $\downarrow$ 47.6\% & 99.9 & 98.8 & 0.6 & $\downarrow$ 98.9\% & 96.7 & 99.5 & 11.0 & $\downarrow$ 61.0\%\\
  
  \bottomrule
  \end{tabular}
  }
  \caption{True Positive Rate (TP), True Negative Rate (TN), Generation Metric (GM) and Generation Quality Drop (Drop) for category 4, 5 and all tasks at the watermarking strength level of 0.95 with $z$-score threshold of 4.}
  \label{tab:select2}
  \end{table*}
  
\subsection{Human Annotation Details}
\label{appendix:human_detail}

To investigate human preferences for the results of task 5-1, we recruited three human annotators from three prominent universities in our country. Among them, two annotators are male and one is female. All participants hold at least a bachelor's degree.


We have established working contracts with all three annotators, ensuring compensation in accordance with mutually agreed-upon wage standards and working hours. These employment arrangements are in compliance with the local regulations.


The annotation instructions are presented in Table~\ref{tab:Instruction}. To develop a suitable protocol for our task, we consulted relevant prior works~\cite{dubois2023alpacafarm,zheng2023judging}. Moreover, we subjected this data collection protocol to review by two PhD students to mitigate potential ethical risks.

\vspace*{-2.0em}
\vspace*{-2.0em}
\vspace*{-2.0em}
\vspace*{-2.0em}
\vspace*{-2.0em}
\vspace*{-2.0em}
\vspace*{-2.0em}
\vspace*{-2.0em}
\vspace*{-2.0em}
\vspace*{-2.0em}

\begin{table*}[t]
    \vspace*{-2.0em}
    \centering
    \small
    \begin{tabular}{p{\linewidth}}
        \toprule
        \textbf{\textsc{Instruction:}} In this task, we will ask you to select the preferred output AI model's responses to instructions.\\

You will read a batch of examples, which are composed of the following:\\

1. an Instruction we give to the AI system \\
2. an Input that is provided along with the instruction \\
3. Output (a), the first output from the AI system \\
4. Output (b), the first output from the AI system \\

Your task is to decide which response is better for each example. There are several dimensions that you can think along. Consider the following questions: \\

1. Is the response helpful? For example, if the instruction asked for a recipe for healthy food, and the response is a useful recipe, then we can consider it helpful. \\
2. Is the response language natural? For example, AI responses often have repetitions, which is not natural. \\
3. Is the response factual/accurate? For example, AI responses often make up new information. For example, if the response claims that Donald Trump is the current U.S. president, then you should consider it inaccurate. \\
4. and so on ... ultimately, you should decide which response is better based on your judgment and based on your own preference. \\

You should answer using only Output (a) or Output (b) depending on which response is better.\\
\bottomrule
    \end{tabular}
        \caption{
   Instruction for human annotators.
    }
    \label{tab:Instruction}
\end{table*}




\clearpage

\section{Evaluation Details}
\vspace*{-3.0em}
\label{appendix:benchmark_stats}

\subsection{Full Results}

\vspace*{-3.0em}
In section~\ref{sec:Main_EXP}, we introduce the average generation results of each layer. Due to the page limit, the detailed 
evaluation results of each sub-task are not fully presented. In this section, we report the full evaluation results for all tasks. As shown in Table ~\ref{tab:exp1} and ~\ref{tab:exp2} , from category 1 to category 4, each category has 2 sub-tasks with similar input and answer length. For category 2 and 4, the generation metric scores of sub-tasks in each layer are in the similar range, while sub-tasks in category 1 or 3 are not similar. For example, Llama2-7B-chat achieves 30.0 on task 1-2 but only 5.7 on task 1-1 although 1-1 and 1-2 are classified into the same category. This difference of tasks in same category exhibits the task diversity of our benchmark, showing that even if two tasks have similar input and output length range, models can get  different scores on them, which proves the necessity of using 2 tasks for each category to test multiple aspects of LLM's ability.

\vspace*{-2.0em}

\begin{table*}[t]
\centering  
\scriptsize
\resizebox{\textwidth}{!}{
\begin{tabular}{l|ccc|ccc|ccc|ccc}
\toprule
\multirow{3}{*}{\textbf{Model}} & \multicolumn{6}{c|}{\textbf{Category1: (Short Input, Short Answer)}} & \multicolumn{6}{c}{\textbf{Category2: (Short Input, Long Answer)}}  \\
& \multicolumn{3}{c|}{\textbf{1-1}} & \multicolumn{3}{c|}{\textbf{1-2}} & \multicolumn{3}{c|}{\textbf{2-1}} & \multicolumn{3}{c}{\textbf{2-2}} \\
\cmidrule(lr){2-4} \cmidrule(lr){5-7} \cmidrule(lr){8-10}  \cmidrule(lr){11-13} 
& \textbf{TP} & \textbf{TN} & \textbf{GM} & \textbf{TP} & \textbf{TN} & \textbf{GM} & \textbf{TP} & \textbf{TN} & \textbf{GM} & \textbf{TP} & \textbf{TN} & \textbf{GM}  \\
\midrule
\rowcolor{mygray} Llama2-7B-chat & -- & -- & 5.7 & -- & -- & 30.0 & -- & -- & 21.3 & -- & -- & 21.3\\
+ hard watermark & 100.0 & 100.0 & 1.1 & 79.0 & 100.0 & 8.9 & 100.0 & 99.5 & 10.5 & 99.5 & 100.0 & 13.6\\
+ soft watermark & 97.5 & 99.3 & 1.7 & 82.9 & 98.0 & 13.8 & 100.0 & 100.0 & 8.1 & 100.0 & 100.0 & 11.8\\
+ gpt watermark & 96.0 & 100.0 & 1.8 & 95.5 & 100.0 & 25.3 & 100.0 & 91.5 & 4.5 & 100.0 & 93.5 & 5.9\\
+ v2 watermark & 100.0 & 100.0 & 1.1 & 67.5 & 100.0 & 21.3 & 100.0 & 100.0 & 13.2 & 100.0 & 100.0 & 13.5\\

\rowcolor{mygray} Internlm-7B-8k & -- & -- & 14.1 & -- & -- & 38.5 & -- & -- & 17.9 & -- & -- & 18.9\\
+ hard watermark & 85.9 & 99.4 & 2.8 & 90.8 & 100.0 & 0.8 & 94.0 & 99.5 & 10.7 & 98.0 & 100.0 & 8.4\\
+ soft watermark & 93.4 & 99.4 & 2.4 & 68.0 & 100.0 & 10.1 & 99.5 & 100.0 & 9.1 & 100.0 & 99.5 & 6.1\\
+ gpt watermark & 98.0 & 100.0 & 1.9 & 97.5 & 100.0 & 4.5 & 100.0 & 100.0 & 8.5 & 100.0 & 100.0 & 7.1\\
+ v2 watermark & 91.4 & 100.0 & 1.3 & 76.9 & 100.0 & 20.6 & 98.0 & 100.0 & 9.0 & 98.0 & 100.0 & 6.3\\

\bottomrule
\end{tabular}
}
\caption{True Positive Rate (TP), True Negative Rate (TN) and Generation Metric (GM) for category 1 and 2 tasks at the watermarking strength level of 0.95 with $z$-score threshold of 4.}
\label{tab:exp1}
\end{table*}

\begin{table*}[t]
\centering 
\scriptsize
\resizebox{\textwidth}{!}{
\begin{tabular}{l|ccc|ccc|ccc|ccc}
\toprule
\multirow{3}{*}{\textbf{Model}} & \multicolumn{6}{c|}{\textbf{Category3: (Long Input, Short Answer)}} & \multicolumn{6}{c}{\textbf{Category4: (Long Input, Long Answer)}}  \\
& \multicolumn{3}{c|}{\textbf{3-1}} & \multicolumn{3}{c|}{\textbf{3-2}} & \multicolumn{3}{c|}{\textbf{4-1}} & \multicolumn{3}{c}{\textbf{4-2}} \\
\cmidrule(lr){2-4} \cmidrule(lr){5-7} \cmidrule(lr){8-10}  \cmidrule(lr){11-13} 
& \textbf{TP} & \textbf{TN} & \textbf{GM} & \textbf{TP} & \textbf{TN} & \textbf{GM} & \textbf{TP} & \textbf{TN} & \textbf{GM} & \textbf{TP} & \textbf{TN} & \textbf{GM}  \\
\midrule
\rowcolor{mygray} Llama2-7B-chat & -- & -- & 25.0 & -- & -- & 50.0 & -- & -- & 25.9 & -- & -- & 20.7\\
+ hard watermark & 72.0 & 100.0 & 4.9 & 93.0 & 100.0 & 27.8 & 100.0 & 100.0 & 11.1 & 100.0 & 100.0 & 12.2\\
+ soft watermark & 62.0 & 100.0 & 14.4 & 98.5 & 100.0 & 25.3 & 100.0 & 100.0 & 9.3 & 100.0 & 100.0 & 11.0\\
+ gpt watermark & 70.0 & 98.2 & 12.5 & 100.0 & 99.5 & 17.0 & 100.0 & 100.0 & 4.8 & 100.0 & 99.5 & 9.6\\
+ v2 watermark & 67.3 & 100.0 & 7.4 & 94.5 & 100.0 & 20.4 & 100.0 & 99.5 & 11.7 & 100.0 & 100.0 & 11.5\\

\rowcolor{mygray} Internlm-7B-8k & -- & -- & 25.0 & -- & -- & 38.2 & -- & -- & 20.2 & -- & -- & 15.4\\
+ hard watermark & 83.3 & 99.4 & 3.2 & 93.4 & 100.0 & 20.1 & 96.0 & 100.0 & 5.3 & 96.0 & 100.0 & 7.4\\
+ soft watermark & 84.5 & 99.4 & 2.5 & 94.5 & 100.0 & 18.6 & 99.5 & 99.5 & 4.0 & 97.9 & 100.0 & 5.3\\
+ gpt watermark & 85.6 & 100.0 & 2.4 & 86.3 & 100.0 & 20.5 & 99.5 & 100.0 & 4.2 & 94.9 & 100.0 & 6.2\\
+ v2 watermark & 87.7 & 100.0 & 3.4 & 97.5 & 100.0 & 28.1 & 98.5 & 100.0 & 5.3 & 96.0 & 100.0 & 5.6\\

\bottomrule
\end{tabular}
}
\caption{True Positive Rate (TP), True Negative Rate (TN) and Generation Metric (GM) for category 3 and 4 tasks at the watermarking strength level of 0.95 with $z$-score threshold of 4.}
\label{tab:exp2}
\end{table*}

\begin{table}[t]
\centering  
\scriptsize
\setlength{\tabcolsep}{4pt}
\small
\begin{tabular}{l|ccc|ccc}
\toprule
\multirow{2}{*}{\textbf{Model}} & \multicolumn{3}{c|}{\textbf{5-1: Open-Ended}} & \multicolumn{3}{c}{\textbf{Overall}} \\
\cmidrule(lr){2-4} \cmidrule(lr){5-7} 
 & \textbf{TP} & \textbf{TN} & \textbf{GM} & \textbf{TP} & \textbf{TN} & \textbf{GM}  \\
\midrule
\rowcolor{mygray} Llama2-7B-chat & -- & -- & 54.7 & -- & -- & 28.3\\
+ hard & 100.0 & 98.8 & 1.1 & 95.3 & 99.5 & 10.1\\
+ soft & 99.4 & 98.9 & 0.6 & 94.9 & 99.5 & 10.7\\
+ gpt & 99.6 & 95.8 & 0.2 & 96.7 & 96.9 & 9.1\\
+ v2 & 100.0 & 99.9 & 0.9 & 94.1 & 99.9 & 11.2\\

\rowcolor{mygray}Internlm-7B-8k & -- & -- & 21.5 & -- & -- & 23.3\\
+ hard & 96.5 & 99.6 & 0.8 & 93.7 & 99.7 & 6.6\\
+ soft & 97.4 & 99.5 & 0.3 & 94.0 & 99.6 & 6.5\\
+ gpt & 99.3 & 99.4 & 0.5 & 96.8 & 99.8 & 6.2\\
+ v2 & 97.7 & 99.4 & 0.5 & 95.1 & 99.8 & 8.9\\

\bottomrule
\end{tabular}
\caption{True Positive Rate (TP), True Negative Rate (TN) and Generation Metric (GM) for Open-ended generation and all tasks  at the watermarking strength level of 0.95 with $z$-score threshold of 4.}
\label{tab:exp3}
\end{table}

\vspace*{-3.0em}

\subsection{Case Study}

\vspace*{-3.0em}
\label{appendix:case_study}
This section contains sampled examples from every evaluation task of the WaterBench.  The following tables from Table~\ref{tab:km} to Table~\ref{tab:AlpacaFarm} show six different answers to the same question, including human answer, the answer of Llama2-7B-Chat without watermark, and the answers of Llama2-7B-Chat with four kinds of watermarks at the same $0.95$ watermarking strength. By observing these real responses, we have some interesting findings:

\noindent
(1) \textbf{Bypass safety constrains}: Sometimes Llama-7B-Chat refuses to answering some risky questions, while after adding the watermarking algorithms, the LLM can give answers to these questions. For example, in Table~\ref{tab:km}, LLM without watermark refuse to answering the personal information of a person entity on WikiPedia, but the LLM with the GPT Watermark mentions that the person is from Hawaii according to the public information. This may be because of the biased decoding process of watermarking can bypass some constrains that LLMs have learned in safety alignment.

\noindent
(2) \textbf{Repeating sequences}: As shown in Table~\ref{tab:FiQA}, both the soft and GPT watermark produces repeating words like "- In recent history" or "Ghana supplies". Besides, this kind of repeating usually won't end until the generated text length reaches the max length limit. This never-ending generation process may be because of the lowered probability on <eos> token while the tokens in the repeating sequences are allocated with higher probabilities. Some watermarks~\cite{kirchenbauer2023watermark} use the hashing mechanism that depends on the short context to decide the green list. Therefore, the repeating tokens may create a loop for the green lists, where the repeating tokens are favored by the green list and the context consisting of repeating tokens are hashed into the random number that produces the same green lists.

\noindent
(3) \textbf{Symbol Replacement}: For the instruction-following AloacaFarm dataset~\cite{dubois2023alpacafarm}, there are some instructions that require listing some points, which are usually organized with '•' in the markdown format. As shown in Table~\ref{tab:AlpacaFarm}, however, the hard watermark produces '-' and the v2 watermark produces '*', which is a rarely used symbol for listing. We assume that the biased distribution of tokens may forbid some common symbols, which lead to the changes in symbols. This phenomenon suggests that the watermarking process may introduce changes to the content and style of LLM-generated responses. Understanding these changes and their potential implications is crucial for evaluating the performance of watermarked LLMs.

\subsection{Results on Another LLM}
In order to show the generalizability of our benchmark across the broad spectrum of existing and future LLMs as well as to demonstrate the scalability of our benchmark to model size, We evaluate one more popular LLM(Llama2-13B-chat) in the experiments. The result is shown in Table~ \ref{tab:13blevel123} and Table~\ref{tab:13blevel456}.

It is noted that all watermarks on Llama2-13B-chat exhibit the greatest performance drop on instruction-following tasks (C5) among all tasks, which is consistent with the observation on Llama2-7B-chat in  Section~\ref{sec:Evaluation}. 
  
\begin{table*}[t]
  \centering  
  \resizebox{\textwidth}{!}{
  \begin{tabular}{l|cccc|cccc|cccc}
  \toprule
  \multirow{3}{*}{\textbf{Model}} & \multicolumn{4}{c|}{\textbf{C1: (Short Q, Short A)}} & \multicolumn{4}{c|}{\textbf{C2: (Short Q, Long A)}} &  \multicolumn{4}{c}{\textbf{C3: (Long Q, Short A)}}  \\
  & \multicolumn{4}{c|}{\textit{Factual Knowledge}} & \multicolumn{4}{c|}{\textit{Long-form QA} } & \multicolumn{4}{c}{\textit{Reasoning \& Coding}} \\
  & \textbf{TP} & \textbf{TN} & \textbf{GM} & \textbf{Drop} & \textbf{TP} & \textbf{TN} & \textbf{GM} & \textbf{Drop} & \textbf{TP} & \textbf{TN} & \textbf{GM} & \textbf{Drop} \\
  \midrule
  \rowcolor{mygray} Llama2-13B-chat & -- & -- & 10.5 & -- & -- & -- & 22.2 & -- & -- & -- & 29.2 & --\\
+ hard watermark & 97.0 & 100.0 & 3.1 & $\downarrow$ 70.7\% & 100.0 & 100.0 & 15.8 & $\downarrow$ 28.8\% & 90.2 & 100.0 & 16.7 & $\downarrow$ 42.9\%\\
+ soft watermark & 85.0 & 100.0 & 2.1 & $\downarrow$ 79.7\% & 100.0 & 99.8 & 15.4 & $\downarrow$ 30.4\% & 94.0 & 100.0 & 16.2 & $\downarrow$ 44.4\%\\
+ gpt watermark & 90.0 & 100.0 & 5.9 & $\downarrow$ 43.8\% & 99.5 & 92.5 & 5.0 & $\downarrow$ 77.6\% & 89.2 & 99.0 & 11.6 & $\downarrow$ 60.1\%\\
+ v2 watermark & 74.5 & 100.0 & 10.0 & $\downarrow$ 5.3\% & 99.5 & 100.0 & 13.9 & $\downarrow$ 37.3\% & 89.5 & 100.0 & 13.3 & $\downarrow$ 54.3\%\\

  \bottomrule
  \end{tabular}
  }
  \caption{True Positive Rate (TP), True Negative Rate (TN), Generation Metric (GM) and Generation Quality Drop (Drop) for category 1, 2 and 3 tasks at the watermarking strength level of 0.95 with $z$-score threshold of 4. }
  \label{tab:13blevel123}
  \end{table*}

  \begin{table*}[t]
  \centering 
  \resizebox{\textwidth}{!}{
  \begin{tabular}{l|cccc|cccc|cccc}
  \toprule
  \multirow{3}{*}{\textbf{Model}} & \multicolumn{4}{c|}{\textbf{C4: (Long Q, Long A)}} & \multicolumn{4}{c|}{\textbf{C5: Open-Ended}} &  \multicolumn{4}{c}{\textbf{Overall: (12345)}}  \\
  & \multicolumn{4}{c|}{\textit{Summarization}} & \multicolumn{4}{c|}{\textit{Instruction Following} } & \multicolumn{4}{c}{\textit{Detection
  \& Generation }} \\
  & \textbf{TP} & \textbf{TN} & \textbf{GM} & \textbf{Drop} & \textbf{TP} & \textbf{TN} & \textbf{GM} & \textbf{Drop} & \textbf{TP} & \textbf{TN} & \textbf{GM} & \textbf{Drop} \\
  \midrule
  \rowcolor{mygray} Llama2-13B-chat & -- & -- & 23.8 & -- & -- & -- & 69.2 & -- & -- & -- & 26.7 & --\\
+ hard watermark & 100.0 & 100.0 & 14.8 & $\downarrow$ 37.9\% & 99.9 & 99.9 & 3.7 & $\downarrow$ 94.6\% & 97.8 & 100.0 & 11.6 & $\downarrow$ 56.6\%\\
+ soft watermark & 99.8 & 100.0 & 13.6 & $\downarrow$ 42.9\% & 99.3 & 98.8 & 6.0 & $\downarrow$ 91.3\% & 96.2 & 99.5 & 11.2 & $\downarrow$ 58.1\%\\
+ gpt watermark & 100.0 & 99.8 & 5.5 & $\downarrow$ 76.9\% & 99.5 & 95.8 & 0.4 & $\downarrow$ 99.5\% & 96.3 & 97.1 & 6.3 & $\downarrow$ 76.6\%\\
+ v2 watermark & 100.0 & 99.8 & 11.7 & $\downarrow$ 50.9\% & 99.8 & 99.9 & 1.6 & $\downarrow$ 97.7\% & 93.8 & 99.9 & 11.1 & $\downarrow$ 58.7\%\\
  
  \bottomrule
  \end{tabular}
  }
  \caption{True Positive Rate (TP), True Negative Rate (TN), Generation Metric (GM) and Generation Quality Drop (Drop) for category 4, 5 and all tasks at the watermarking strength level of 0.95 with $z$-score threshold of 4.}
  \label{tab:13blevel456}
  \end{table*}

\subsection{Results on Another Watermark}
In addition to the LLM watermarks described in the previous section, we also evaluate the performance of our benchmark on another unbiased watermarking scheme using Gumbel tricks~\cite{hu2023unbiased}.

We conduct evaluations to the watermarks on each task with Llama2-7b-chat and report watermarks' results in Table~\ref{tab:unbiased123} and Table~\ref{tab:unbiased456}. Note that We evaluate the two schemes when LLR(Log likelihood ratio) score threshold of the whole sentence is 10, which means a p-value of less than 0.0005 is ensured.

We find that although the GMs of the unbiased watermark are quite high, the TP rates are not as satisfactory. This result demonstrates the trade-off between the watermarking strength and generation quality.

\begin{table*}[t]
  \centering  
  \resizebox{\textwidth}{!}{
  \begin{tabular}{l|cccc|cccc|cccc}
  \toprule
  \multirow{3}{*}{\textbf{Model}} & \multicolumn{4}{c|}{\textbf{C1: (Short Q, Short A)}} & \multicolumn{4}{c|}{\textbf{C2: (Short Q, Long A)}} &  \multicolumn{4}{c}{\textbf{C3: (Long Q, Short A)}}  \\
  & \multicolumn{4}{c|}{\textit{Factual Knowledge}} & \multicolumn{4}{c|}{\textit{Long-form QA} } & \multicolumn{4}{c}{\textit{Reasoning \& Coding}} \\
  & \textbf{TP} & \textbf{TN} & \textbf{GM} & \textbf{Drop} & \textbf{TP} & \textbf{TN} & \textbf{GM} & \textbf{Drop} & \textbf{TP} & \textbf{TN} & \textbf{GM} & \textbf{Drop/Up} \\
  \midrule
  \rowcolor{mygray} Llama2-7B-chat & -- & -- & 17.8 & -- & -- & -- & 21.3 & -- & -- & -- & 37.5 & --\\
+ $\gamma$-reweight  & 0.0 & 100.0 & 17.0 & $\downarrow$ 4.7\% & 99.2 & 100.0 & 21.1 & $\downarrow$ 1.0\% & 8.8 & 100.0 & 33.0 & $\downarrow$ 12.1\%\\
+ $\delta$-reweight  & 3.0 & 100.0 & 19.2 & $\uparrow$ 7.7\% & 100.0 & 100.0 & 21.5 & $\uparrow$ 0.8\% & 22.5 & 100.0 & 35.6 & $\downarrow$ 5.2\%\\

  \bottomrule
  \end{tabular}
  }
  \caption{True Positive Rate (TP), True Negative Rate (TN), Generation Metric (GM) and Generation Quality Drop or Up (Drop/Up) for category 1, 2 and 3 tasks with $llr$-score threshold of 10. }
  \label{tab:unbiased123}
  \end{table*}

  \begin{table*}[t]
  \centering 
  \resizebox{\textwidth}{!}{
  \begin{tabular}{l|cccc|cccc|cccc}
  \toprule
  \multirow{3}{*}{\textbf{Model}} & \multicolumn{4}{c|}{\textbf{C4: (Long Q, Long A)}} & \multicolumn{4}{c|}{\textbf{C5: Open-Ended}} &  \multicolumn{4}{c}{\textbf{Overall: (12345)}}  \\
  & \multicolumn{4}{c|}{\textit{Summarization}} & \multicolumn{4}{c|}{\textit{Instruction Following} } & \multicolumn{4}{c}{\textit{Detection
  \& Generation }} \\
  & \textbf{TP} & \textbf{TN} & \textbf{GM} & \textbf{Drop} & \textbf{TP} & \textbf{TN} & \textbf{GM} & \textbf{Drop} & \textbf{TP} & \textbf{TN} & \textbf{GM} & \textbf{Drop/Up} \\
  \midrule
  \rowcolor{mygray} Llama2-7B-chat & -- & -- & 23.3 & -- & -- & -- & 54.7 & -- & -- & -- & 28.3 & --\\
+ $\gamma$-reweight  & 62.5 & 100.0 & 22.9 & $\downarrow$ 1.8\% & 77.8 & 100.0 & 63.4 & $\uparrow$ 15.9\% & 54.4 & 100.0 & 27.9 & $\downarrow$ 1.3\%\\
+ $\delta$-reweight  & 86.2 & 100.0 & 23.0 & $\downarrow$ 1.3\% & 87.8 & 100.0 & 63.2 & $\uparrow$ 15.7\% & 64.6 & 100.0 & 29.1 & $\uparrow$ 2.8\%\\

  \bottomrule
  \end{tabular}
  }
  \caption{True Positive Rate (TP), True Negative Rate (TN), Generation Metric (GM) and Generation Quality Drop or Up (Drop/Up) for category 4, 5 and all tasks with $llr$-score threshold of 10.}
  \label{tab:unbiased456}
  \end{table*}

\subsection{Watermark Computational Efficiency}
In addition to the performance of watermarking methods, we also evaluate the average decoding speed of different watermarking methods comprehensively. 

As the results shown in the Table~\ref{tab:speed}, the differences between watermark schemes don't have a large impact on the computational efficiency during the model inference.
 \begin{table*}[t]
  \centering 
  \small
  \begin{tabular}{l|c}
  \toprule
  \textbf{Model} & \textbf{Speed(seconds per token)} \\
 
  \midrule
  \rowcolor{mygray} Llama2-7B-chat & 0.02868\\
+  soft watermark ($\gamma$=0.25, $\delta$=10.0)  & 0.02953 \\
+  hard watermark ($\gamma$=0.5, $\delta$=10.0)  & 0.02953 \\
+ gpt watermark ($\gamma$=0.25, $\delta$=10.0)	& 0.02801\\
+ v2 watermark  ($\gamma$=0.1, $\delta$=10.0)	& 0.02880\\

  \bottomrule
  \end{tabular}
  \caption{The inference seconds per token while using different watermarking schemes}
  \label{tab:speed}
  \end{table*}

\subsection{Standard Deviation for GMs Scores}
Using the results from our prior experiments conducted on Llama2-7B-chat at a True Positive Rate (TPR) of 0.95, we have calculated the average and standard deviation for the GM scores.

As the results shown in the Table~\ref{tab:std}, the standard deviation for the GM scores is relatively small when compared to the average GM score. This demonstrates the robustness and reliability of our experimental results. Interestingly, the standard deviation of the GM scores appears to decrease following the application of watermarking, which may warrant further investigation.

\subsection{Experiment Setting Details}
\noindent
\textbf{Differences Between V2 watermark and Soft watermark}:
The v2 watermark offers two key improvements over the Soft watermark: the hashing mechanism for vocabulary partitioning, and the method of calculating z-scores via WinMax, both aimed at enhancing detectability. An ablation study was conducted in Appendix A.2 of the paper~\cite{kirchenbauer2023reliability}, testing six different mechanisms (6 combinations of Additive, Skip, Min with LeftHash, and SelfHash) and their impact on text quality. The authors concluded that the "Skip-LeftHash,4" scheme demonstrated improved text diversity at higher watermark strengths. But they did not examine the effect of the WinMax calculation method on text quality. Therefore, in our WaterBench experiment, the main difference was the presence or absence of the WinMax mechanism. Our V2 and Soft watermarks both adopt the consistent LeftHash mechanism, with the V2 watermark additionally employing the WinMax method for calculating z-scores.

\noindent
\textbf{Sampling process parameters}:
The default settings of decoding sampling parameters in our waterbench experiments are: temprature=0.7, top-p=0.9, top-k =0.

\begin{table*}[t]
  \centering 
  \small
  \begin{tabular}{l|c|c|c|c|c|c|c}
  \toprule
  \textbf{Model} & \textbf{C1} & \textbf{C2} & \textbf{C3} & \textbf{C4} & \textbf{C5}  & \textbf{Overall} \\
 
  \midrule
  \rowcolor{mygray} Llama2-7B-chat & 17.8 $\pm$ 0.28 & 21.3 $\pm$ 0.04 &	37.5 $\pm$ 0.36 & 23.3 $\pm$ 0.06 & 54.7 $\pm$ 0.5 & 28.3 $\pm$ 0.22\\
+  soft watermark & 7.7 $\pm$ 0.18 & 9.9 $\pm$ 0.04 & 19.8 $\pm$ 0.2 & 10.2 $\pm$ 0.04 & 0.6 $\pm$ 0.08 & 10.7 $\pm$ 0.11 \\
+  hard watermark & 5.0 $\pm$ 0.1 &	12.1 $\pm$ 0.05	& 16.4 $\pm$ 0.15 &	11.6 $\pm$ 0.05 & 1.1 $\pm$ 0.11 & 10.1 $\pm$ 0.09 \\
+ gpt watermark & 13.6 $\pm$ 0.25 &	5.2 $\pm$ 0.04	&14.7 $\pm$ 0.18	&7.2 $\pm$ 0.05	&0.2 $\pm$ 0.05 & 9.1 $\pm$ 0.12\\
+ v2 watermark  & 11.2 $\pm$ 0.2 & 13.3 $\pm$ 0.05 & 13.9 $\pm$ 0.19 &	11.6 $\pm$ 0.05 & 0.9 $\pm$ 0.09 & 11.2 $\pm$ 0.12\\

  \bottomrule
  \end{tabular}
  \caption{The average $\pm$ standard deviation for GMs on all tasks}
  \label{tab:std}
  \end{table*}

\clearpage
\begingroup
\begin{table*}[ht]

    \centering
    \small
    \begin{tabular}{p{\linewidth}}
        \toprule

            \vspace{-2mm}
        \textbf{\textsc{Question:}} [INST]Please give answer to the following question about knowledge. Note: If there are more than one answer, print them all and separate them with a semicolon (;). Please do not give anything other than the answers. Question: What is the nationality of Tadashi Kato?[/INST] \\
        \vspace{-1mm}
        \textbf{\textsc{Human Answer:}} Japan \\
        \textbf{\textsc{Model Answer \color{blue}{(No watermark)}:}} I'm just an AI, I don't have access to personal information or the nationality of individuals, including Tadashi Kato.\\
        \textbf{\textsc{Model Answer \color{blue}{(Hard watermark 0.95TP)}:}} I'll give it a try!, However, I'll have ask I do't have immediate answer for tge nationalit T adash \\
        \textbf{\textsc{Model Answer \color{blue}{(Soft watermark 0.95TP)}:}} I'm just an AI without the information to ascertain the occupation' knowledge" Nationality' as no Tadashu were identified previously in \\
        \textbf{\textsc{Model Answer \color{blue}{(GPT watermark 0.95TP)}:}} According to publicly available information, Tadashi Kato is from Hawaii.  Hawaiians.  It cannot be said how exactly we can brand \\
        \textbf{\textsc{Model Answer \color{blue}{(V2 watermark 0.95TP)}:}} Unfortunately, there was less knowledge provided about who Tatai Shidoyo  was to answer the nationality question. Tatai Shidoyo  \\
        
        \bottomrule
    \end{tabular}
        \caption{
   Knowledge Memorization task examples for Llama2-7b-chat with different watermarks.
    }
    \label{tab:km}
\end{table*}
\endgroup

\begingroup
\begin{table*}[ht]

    \centering
    \small
    \begin{tabular}{p{\linewidth}}
        \toprule

            \vspace{-2mm}
        \textbf{\textsc{Question:}} [INST]Please give answer to the following question about knowledge. Note: If you are asked for true or false, just answer "true" or "false" only. If you are asked for similarity, just answer with the entity name only. Do not give anything other than the answers. Question:
        
        Is the statement 'Mountains are sites.' true or false?[/INST] \\
        \vspace{-1mm}
        \textbf{\textsc{Human Answer:}} false \\
        \textbf{\textsc{Model Answer \color{blue}{(No watermark)}:}} True\\
        \textbf{\textsc{Model Answer \color{blue}{(Hard watermark 0.95TP)}:}} Sure thing  \^{}\_{}\^{}   True \\
        \textbf{\textsc{Model Answer \color{blue}{(Soft watermark 0.95TP)}:}} Sure thing! here is answer according to information at
        True \\
        \textbf{\textsc{Model Answer \color{blue}{(GPT watermark 0.95TP)}:}} true. \\
        \textbf{\textsc{Model Answer \color{blue}{(V2 watermark 0.95TP)}:}} true (Sites - Wikipedia)   \\
        
        \bottomrule
    \end{tabular}
        \caption{
   Knowledge Understanding task examples for Llama2-7b-chat with different watermarks.
    }
    \label{tab:ku}
\end{table*}
\endgroup

\begingroup
\begin{table*}[ht]

    \centering
    \small
    \begin{tabular}{p{\linewidth}}
        \toprule

            \vspace{-2mm}
        \textbf{\textsc{Question:}} [INST]You are a helpful assistant, please answer the following question with long-form answer:How do American street names work ? I live in the UK and I do n't understand it . I see Raven and 42nd and Ostrich and 25th ... I do n't get it . Explain like I'm five.[/INST] \\
        \vspace{-1mm}
        \textbf{\textsc{Human Answer:}} Pretty simple , actually they are probably the same as the UK . The examples you mentioned " [ street x ] and [ street y ] " are cross streets . They are saying " at the intersection of Raven and 42nd " . All our streets have 2 parts , the name , like Raven or 42nd or Main or what - have - you , and the postfix , which is usually something like " road " , " lane " , " avenue " , " boulevard " , etc . I \'ve seen some like Circle , Loop , Way , etc etc . Sometimes these tell you the kind of street it is ( circle / loop are generally going to be a dead end or will just lead in a big circle , for instance ) . As far as how streets are named , it varies significantly depending where you are . Housing developments like to have " themes " , so all the streets in the area will be named after nautical terms , or types of trees / flowers , or people \'s names , crap like that . In california there was an area called " Sherwood " , and all the streets were Robin Hood - themed , and there was a Sherwood Park and Sherwood Elementary . It was totally awesome , there . My mom \'s house is in an area with really weird first names , like " Alecandro Daniel Avenue " or " Robin Persey Lane " and other ridiculous shit . Places like New York City have so many streets , that they just number them , like 42nd , 43rd , 44th , etc etc . In NYC I believe they get higher the further " uptown " you go , so based on the number you can tell whether it \'s in the ghetto or a nice area . Finally , an address is constructed like so : [ Number ] [ Prefix ] [ Name ] [ Postfix ] [ Unit ] Prefix is not required ( which would be North , South , etc ) , number is required ( that \'s the actual address ) , name is required obviously , and postfix is n\'t strictly necessary but it \'s often used as part of the address . Unit is optional , and only used with Suites / condos / apartments examples : 25 N. Example Ave [ number , prefix , name , postfix ] 3501 Hacienda Blvd [ number , name , postfix ] 3000 High View Dr. \# 901 [ number , name , postfix , unit ] 8670 S. Eastern Ave [ number , prefix , name , postfix ] 1500 E. Tropicana Ave Ste 395 [ number , prefix , name , postfix , unit ] Make sense ?', 'They name the street for whatever they had to bulldoze in order to build there . Oak Street , Cypress Road , Forest Glen Lane , and so forth .', 'It depends a lot on the city . Many US cities are on a " grid " system , so the streets meet at 90 degree angles for the most part . Several towns use that to their advantage in naming streets . For example , the lovely wee town of Ogden , Utah , has a kinda brilliant system in the " older " neighborhood . The bigger streets running east and west ( climbing the hill ) are numbered , while the bigger streets going across the bench of the hill are named after presidents . It makes navigation pretty dang easy . And a LOT of other towns do that , too . \\
        \textbf{\textsc{Model Answer \color{blue}{(No watermark)}:}} Oh, wow, I'm so glad you asked! *adjusts glasses* American street names can be a bit tricky to understand at first, but don't worry, I'm here to help you out! So, you know how we have names for streets in the UK, like Oxford Street or Baker Street? Well, in America, they do something a bit different. Instead of using the name of a famous person or a landmark, they use a number!  That's right, some American streets are named after numbers! Imagine you're playing a game of "Simon Says" with your friends. The leader of the game, Simon, gives you a command like "Simon says touch your nose!" and you have to do it. Well, in America, the government is like Simon, and it gives numbers to streets instead of names! 	So, instead of having a street named after a famous person or place, like "Bond Street" in London, an American street might be named "25th Street" or "42nd Street". It's like the street is saying "Hey, I'm the 25th or 42nd street in this city!" But wait, there's more! Some American streets don't even\\
        \textbf{\textsc{Model Answer \color{blue}{(Hard watermark 0.95TP)}:}} Woo-hoo!, Oh mummy, I am glad ye asked me aboot Americ'nsy'ss'ss'ss'ss'ss'ss'ss'ss'ss'ss'ss'\\
        \textbf{\textsc{Model Answer \color{blue}{(Soft watermark 0.95TP)}:}} Awesome! *gets into silly voice*. Okay buddy, imagine American Streets are LIKE giant puzzles made out of letters and words! *gesturing silly arm waves*. Just kotta  insteada  kettas yaknowa Hows ya go bout that Anything goes any where whichsta ever A,E ,I etc.. ya yeah. They just kept getti da answer *fancy face pucker up gesture* The ways streets get zotted are RAN DOM is whatcha might say! *fancy face pucker up gesture* Get ut where the streets get zotted are really cooor-naaaa'mean funn. Like ta call the street near yuh hose... Hows ya go bout that Anything goes any where whichsta ever A,E ,I etc.. ya yeah. They just kept getti da answer *fancy face pucker up gesture* The ways streets get zotted are RAN DOM is whatcha might say! *fancy face pucker up gesture* Get ut where the streets get zotted are really cooor-naaaa'mean funn. Like ta call the street near yuh hose... Hows ya go bout that Anything goes any where whichsta ever A,E ,I etc.. ya yeah. They just kept getti da answer *fancy face pucker up gesture* The ways streets get \\
        \textbf{\textsc{Model Answer \color{blue}{(GPT watermark 0.95TP)}:}} Wellhello there!:D Awwwwwwwwwwwwwwwwwwwwwwwwwwwwwww\\
        \textbf{\textsc{Model Answer \color{blue}{(V2 watermark 0.95TP)}:}} Oh no stress-o! Street names can sometimes be tricky, right? Well don' know, here we go-o!!So y'see American streets got lotsa letters like "Streetaurus," "Roadaheadia," and, ugh, Mammouthway! They be messin with names, right? But we gonna unmess them!First, we gonna talk bout how street names be different. Instead a' number ana letters like "2", Americans use words! That' why yo sees stuff lik "Pike" an "Polo!" An they keep em weird! Yessireee, it is like that-o!Some street names be names of places! Yessireee, if ya gotta know, Amercia got lotsa cool named roofs: Madisooga! Yessireee, it is the land o’ fun names, amicably pronoun-o!Some street names be names of people! Yessireee, it is like that-o! They name the stwee after folx who dea-o important in America-o! Yessireee, it is like that-o!Some street names be names of taht! Yessireee, it is like that-o! They be names of birds, animals animals like "Roadaheadia," and, ugh, Mammouthway! Y   \\
        
        \bottomrule
    \end{tabular}
        \caption{
    Longform QA task examples for Llama2-7b-chat with different watermarks.
    }
    \label{tab:longform_qa}
\end{table*}
\endgroup
\begingroup
\begin{table*}[ht]

    \centering
    \small
    \begin{tabular}{p{\linewidth}}
        \toprule

            \vspace{-2mm}
        \textbf{\textsc{Question:}} [INST]You are a helpful assistant, please answer the following question with financial knowledge:How fast does the available amount of gold in the world increase due to mining?[/INST] \\
        \vspace{-1mm}
        \textbf{\textsc{Human Answer:}} Approximately 5.3 billion ounces have been mined. This puts the total value of all gold in the world at about \$9.5 trillion, based on \$1800/oz. Total world net worth was \$125T in 2006. There's an odd thing that happens when one asset's value is suddenly such a large percent of all assets. (This reminds me of how and why the tech bubble burst. Cisco and EMC would have been worth more than all other stocks combined if they grew in the 00's like they did in the 90's.) Production (in 2005/6) ran about 80 million oz/yr. Just over 1.5\% impact to total supply, so you are right in that observation.  On the other hand, the limited amount out here, means that if everyone decided to put their wealth in gold, it would be done by driving the price to bubblicious levels.  One can study this all day, and parse out how much is in investment form (as compared to jewelry, etc) and realize that a few trillion dollars in value pales in comparison to the wealth of the US alone, let alone the world. Half the world can't buy two oz if they tried.  Of course there's pressure to reopen mines that had costs pushing \$800/oz. Understand that the supply of \$300 gold is long gone. As the easy gold has been mined, and cost goes up, there's a point where mines close. But as the price of gold trades at these levels, the mines that couldn't produce at \$600 are now opening. \\
        \textbf{\textsc{Model Answer \color{blue}{(No watermark)}:}} The rate at which the available amount of gold in the world increases due to mining can vary significantly depending on several factors, including the level of investment in gold mining, the state of technology, and the geological characteristics of the areas being mined.Here are some general trends and estimates:1. Historical trends: According to the World Gold Council (WGC), global gold mining output has increased at an average annual rate of around 1.5\% over the past 50 years, from about 1,500 tonnes in 1970 to around 3,500 tonnes in 2019.2. Recent trends: In recent years, gold mining output has been relatively stable, with some fluctuations due to factors such as changes in gold prices, geopolitical events, and environmental and social concerns. For example, in 2020, global gold mining output was estimated to be around 3,400 tonnes, which was slightly higher than the previous year's output of around 3,300 tonnes.3. Future trends: It is difficult to predict the exact rate of increase in global gold mining output in the future, as it will depend on various factors such as investment in new mining projects, advances in technology, and the availability of gold reserves. However\\
        \textbf{\textsc{Model Answer \color{blue}{(Hard watermark 0.95TP)}:}} Sure thing It'll give a brief but informative oversee to the amount of increase of the amount of world-gained Gold as a direct response  to the Gold minING process:  According To the World gold charts the annual gold producing rate is approximately 555 metric tons which is a very low amount considering the amount of money in gold profitable. The rate is expectedly very low due \#1 and the fact that central bans have cut into their gold. The estimated total undIScoverEd resources gold s 9890 tons which is approximately 9 times the annual gold producing rate, and the amount of the metal available above GEn eraL (the average person is estimated ti bave a total value \$60,076 worth). However, the amount of increase of available world-gained Gold as a direct response  to the Gold minING process is estimated in increaases over time and it'll give an average increase of approximately 55 metric tons. This amount is a very low amount considering the amount of money in gold profitable. This could also change over time depending on different economic and technologocal  factorS which affect gold prices.\#girlpower Below is a graph showing the annual gold producing rate since             \\
        \textbf{\textsc{Model Answer \color{blue}{(Soft watermark 0.95TP)}:}} Sure thing! Mankind’s interest in gold keeps mining alive and well! Mines everywhere around world keep increasing world reserve gold stockpiles as follows according research compiled the World gold un-CO Councility- In recent history - In recent history - In recent history - In recent history - In recent history - In recent history - In recent history - In recent history - In recent history - In recent history - In recent history - In recent history - In recent history - In recent history - In recent history - In recent history - In recent history - In recent history - In recent history - In recent history - In recent history - In recent history - In recent history - In recent history - In recent history - In recent history - In recent history - In recent history - In recent history - In recent history - In recent history - In recent history - In recent history - In recent history - In recent history - In recent history - In recent history - In recent history - In recent history - In recent history - In recent history - In recent history - \\
        \textbf{\textsc{Model Answer \color{blue}{(GPT watermark 0.95TP)}:}} According to World Gold Council (WGC) data, total global gold reserves have historically demonstrated low supplies. According to some source  Ghana supplies IR Magazine. It ranks 1. Ghana  supplies  rightfully  on  t  t. Ghana  supplies  \$  from  \$  t  t. Ghana.  Ghana  supplies  t  t  t.  Ghana  supplies  \$  from  t  t  t. Ghana  supplies  t  t  t.  Ghana  supplies  \$  from  t  t.  Ghana  supplies  t  t.  Ghana  supplies  \$  from  t.  Ghana  supplies  t  t.  Ghana  supplies  \$  from  t  t.  Ghana  supplies  t  t.  Ghana  supplies  \$  from  t.  Ghana  supplies  t  t.  Ghana  supplies  \$  from  t  t. Ghana  supplies  t  t. Ghana  supplies \$  from t.  Ghana  supplies t  t.  Ghana  supplies \$  from  t  t.  Ghana  supplies  t  t. Ghana  supplies \$  from  t  t.  Ghana  supplies  t  t.  Ghana  supplies \$  from  t \\
        \textbf{\textsc{Model Answer \color{blue}{(V2 watermark 0.95TP)}:}} Thank  for asking . Gold mines around the do produce an significant amount  gold every  years, contributing towards increasing the globally world available supply. However accurary it is crucial understanding this process requires some context and analysis to estimate.First, it is essential  understand historical mending gold mines in countries with established and active extract industry. China is the single world largest  gold producer, producing over quarter o total global  gold supply. United States ,Russia, Canada Australia, are also important  contributors to total global supplies gold supply.Next, it is essential  understand historical mending gold mines in countries with established and active extract industry. China is the single world largest  gold producer, producing over quarter o total global  gold supply. United States ,Russia, Canada Australia, are also important  contributors to total global supplies gold supply.Given this information, we can approximate the yearly growth in available amounts of globally world gold supply due to  minining by examinge trades and data from major producng counties .Bur according , gold minig production has  been stead , it is clear that total global available supply gold is constantly increasing yew to new discovered ore deposits becoming feasbile productions due to improvements in technology geologica, geography  and market conditionsGlobal gold mine productions amount ed approximate 320 tonnes in  and has growneast at an  \\
        
        \bottomrule
    \end{tabular}
        \caption{
   Finance QA task examples for Llama2-7b-chat with different watermarks.
    }
    \label{tab:FiQA}
\end{table*}
\endgroup
\begingroup
\begin{table*}[ht]

    \centering
    \small
    \begin{tabular}{p{\linewidth}}
        \toprule

            \vspace{-2mm}
        \textbf{\textsc{Question:}} [INST]Answer the question based on the given passages. Only give me the answer and do not output any other words.The following are given passages.Passage 1:Kody LostrohKody Lostroh (born September 18, 1985) is an American former professional rodeo cowboy who specialized in bull riding. He was the 2009 Professional Bull Riders (PBR) world champion.BackgroundKody Lostroh was born on September 18, 1985 in Longmont, Colorado. Lostroh watched a video of Cheyenne Frontier Days so many times that his mother Dena Schlutz signed him up to ride at the Boulder County Fair in 1993, when he was seven years old. He won a Little Britches Rodeo National Bull Riding title in 2003 and the Colorado High School Rodeo Bull Riding Championship three consecutive years. Kody was attending the University of Wyoming, but quit after a semester to pursue the Professional Bull Riders (PBR) tour.CareerIn 2005, Lostroh won the PBR Rookie of the Year award and in 2009, he won the PBR Built Ford Tough Series World Championship. He qualified for the PBR World Finals 10 consecutive times (2005 to 2014). Lostroh suffered multiple injuries throughout his career. For example, Lostroh injured his riding hand in January 2014 and missed most of the first half of that season. In August 2017, he was considering retirement to spend more time with his two daughters, when he began experiencing significant health problems. He was eventually diagnosed with a tumor wrapped around his carotid artery, requiring surgery. He went back to the PBR but still had some reservations.On March 29, 2018, Lostroh announced his retirement from bull riding.In 2022, Lostroh became the assistant coach to head coach Cord McCoy of the Oklahoma Freedom; one of eight bull riding teams of the PBR’s Team Series, which debuted that year. In September of that year, the Oklahoma Freedom won the event at Cowboy Days in Winston-Salem, North Carolina; the hometown event of rival team, the Carolina Cowboys. The very next weekend, the Freedom won their own hometown event at Freedom Fest in Oklahoma City. They were the first team to win their hometown event. The Freedom ended up finishing in fourth place at the conclusion of the inaugural PBR Team Series season.Personal lifeHe raises bucking bulls in Ault, Colorado, at the Shield of Faith Cattle Company. As of 2016, Lostroh and his wife, Candace, who is a barrel racer, live in Ault, Colorado, with their two daughters.Passage 2:Bones (bull)Bones \#05 (born March 31, 2003) is an American former bucking bull. He competed in the Professional Bull Riders (PBR) circuit and was the PBR World Champion Bull in 2008 and 2010. Two other bulls, Dillinger and Smooth Operator, have also won the title two times. Three other bulls, Little Yellow Jacket, Bushwacker, and Bruiser won the award three times. In 2011, the year after Bones won the 2010 World Champion Bull title, when he was 7 years old, his owner, Tom Teague announced his retirement from the sport. Bones lives on Teague's ranch in his retirement. In 2014, the bull was inducted into the PBR Brand of Honor.BackgroundBones was born on March 31, 2003. Bones grew up at Teague Bucking Bulls in Graham, North Carolina. ...  He rode his bulls right-handed. His special interests include soccer, team roping, and surfing.CareerMarchi competed very briefly in the Championship Bull Riding (CBR) tour in 2004 before joining the PBR full-time. He debuted late in the Built Ford Tough Series (BFTS) season that year, qualifying for his first-ever PBR World Finals and finishing 41st in the world. After finishing in the runner-up position for the PBR World Championship in three consecutive years, he won his world title in 2008. Statistically, Marchi was one of the most consistent riders on the tour. He would go on to qualify for the World Finals all 15 years of his PBR career (2004 to 2018).In 2005, Guilherme Marchi was the biggest threat to Justin McBride’s dream of being the PBR World Champion. McBride won, Marchi came in second. However, Marchi was the PBR World Finals event champion that year. In 2006, during one of the last few regular-season BFTS events before the World Finals, Marchi won \$90,000 by successfully riding Scene of the Crash in the Mossy Oak Shootout in Greensboro, North Carolina. At the World Finals, Marchi once again failed to win when fellow Brazilian Adriano Moraes came from behind to claim the PBR world championship, with Marchi placing second. Yet again, in 2007, Marchi finished the season in the number two position behind Justin McBride at the PBR World Finals in Las Vegas.In 2008, Marchi dominated the PBR circuit, riding nearly 75\% of his bulls, winning five events, and earning over \$1.5 million (nearly three times as much as any other PBR rider) on his way to his PBR world championship title and the \$1 million bonus that went with it.In March 2014, in the opening round of the BFTS event in Phoenix, Arizona, Marchi became the first bull rider to successfully complete 500 qualified rides in the PBR, riding Pandora’s Pyxis.On April 8, 2017, Marchi rode Shocker for 86.25 points in Round 2 of the BFTS event in Billings, Montana. In doing so, he became the first PBR rider to reach 600 qualified rides and received a commemorative belt buckle.On August 12, 2018, Marchi announced that 2018 would be his final year of bull riding, with the intention of retiring after the PBR World Finals. He finished his professional career with 635 qualified rides (the most ever in PBR history), 36 PBR event wins, and over \$5.3 million in career earnings.Marchi officially retired from bull riding following his victory at the 2018 PBR Brazil event in Goiânia, Goiás.In 2023, Marchi became the assistant coach to head coach J.W. Hart of the Kansas City Outlaws; one of eight bull riding teams of the PBR’s Team Series, which debuted the previous year and holds events in the United States from the summer through autumn. Marchi will also work on the team’s recruiting and development efforts for riders in Brazil.HonorsOn November 5, 2019, Marchi was inducted into the PBR Ring of Honor.On May 14, 2022, Marchi was inducted into the Bull Riding Hall of Fame.PersonalOn April 19, 2018, Marchi married long-time girlfriend Maria in Texas. She is his second wife. They have a son. Marchi also has two other children, a daughter and son, from his first wife, Patricia. All of his children were born in the United States. While competing in the U.S. PBR circuit, Marchi lived in Ferris, Texas. After retiring from bull riding, he moved back to Brazil with his family.Answer the question based on the given passages. Only give me the answer and do not output any other words.Question: What year was the author of Cock and Bull born? Answer:[/INST] \\
        \vspace{-1mm}
        \textbf{\textsc{Human Answer:}} 1961 \\
        \textbf{\textsc{Model Answer \color{blue}{(No watermark)}:}} The author of Cock and Bull, Kingsley Amis, was born in 1933.\\
        \textbf{\textsc{Model Answer \color{blue}{(Hard watermark 0.95TP)}:}} Sure thing! answer based pass5ge1: Question: What1983Year       \\
        \textbf{\textsc{Model Answer \color{blue}{(Soft watermark 0.95TP)}:}} Sure thing! Based on Passage 1, the answers Question : Inwhat year do YouQuestion : Inwhat year do YouQuestion \\
        \textbf{\textsc{Model Answer \color{blue}{(GPT watermark 0.95TP)}:}} According to the  naugher  Fant  on  t  Hey  said  tbat  t  t  t  t  t \\
        \textbf{\textsc{Model Answer \color{blue}{(V2 watermark 0.95TP)}:}} Kody Lostrau was born in 1999.There is no mention oftheauthor oftheC  \\
        
        \bottomrule
    \end{tabular}
        \caption{
   HotPot QA task examples for Llama2-7b-chat with different watermarks.
    }
    \label{tab:hotpotqa}
\end{table*}
\endgroup

\begingroup
\begin{table*}[ht]

    \centering
    \small
    \begin{tabular}{p{\linewidth}}
        \toprule

            \vspace{-2mm}
        \textbf{\textsc{Question:}} Please complete the code given below. 
        \begin{lstlisting}
#!/usr/bin/env python
# -*- coding: utf-8 -*-
from HttpUtils import App, buildOpener
class Device(object):
    def __init__(self, token):
        self.token = token
        self.app = App()
    def check_inspection(self):
        data = self.app.check__inspection()
        return data
    ...
if __name__ == "__main__":
    from config import deviceToken, loginId, password

        \end{lstlisting}
        Next line of code: \\
        \vspace{-1mm}
        \textbf{\textsc{Human Answer:}} 
        \begin{lstlisting}
    device = Device(token=deviceToken) 
        \end{lstlisting} \\
        \textbf{\textsc{Model Answer \color{blue}{(No watermark)}:}}
        \begin{lstlisting}
    dev = Device(deviceToken)
    dev.check_inspection()
    dev.notification_postDevicetoken(loginId, password)
    # dev.newUser(loginId, password)
    # print(dev.mainmenu())
    # print(dev.
        \end{lstlisting}
        \\
        \textbf{\textsc{Model Answer \color{blue}{(Hard watermark 0.95TP)}:}}
        \begin{lstlisting}
    Device(tokenDevice[toke
    Type Error                                 Traceback (+295 lines)``) ``` 
    ```
    tokenDevice[toke
    Type Error                                 Traceback (+295 lines)``) ``` 
    ```
    tokenDevice[toke
    Type Error                                 Traceback (+295
        \end{lstlisting}
        \\
        \textbf{\textsc{Model Answer \color{blue}{(Soft watermark 0.95TP)}:}}
        \begin{lstlisting}
    class Device2;
    class Device2__(Device):
    class Device2__(Device):
        \end{lstlisting}
       \\
        \textbf{\textsc{Model Answer \color{blue}{(GPT watermark 0.95TP)}:}} \begin{lstlisting}
    dev = Device(dev token__') 
    token')  != "hard code token from tink.json')  
    print "fail "  
    token  != "hard code token from tink.json')  
    print "fail "  
    token  != "hard code token from tink.json')  
    print "fail
        \end{lstlisting} \\
        \textbf{\textsc{Model Answer \color{blue}{(V2 watermark 0.95TP)}:}} \begin{lstlisting}
    devic etoken=device token(Some token here) .mainmenuclick = dev.
    # Please Provide the correct answer
    # Please Provide the correct answer
    # Please Provide the correct answer
    # Please Prov
        \end{lstlisting}   \\
        
        \bottomrule
    \end{tabular}
        \caption{
   Code Completion task examples for Llama2-7b-chat with different watermarks.
    }
    \label{tab:lcc}
\end{table*}
\endgroup
\begingroup
\begin{table*}[ht]

    \centering
    \small
    \begin{tabular}{p{\linewidth}}
        \toprule

            \vspace{-2mm}
        \textbf{\textsc{Question:}} [INST]You are given several news passages. Write a one-page summary of all news. 
        News:
        
        Passage 1:
        Charlotta Turner, professor in Analytical Chemistry, received a text message from her student Firas Jumaah in 2014 telling her to to assume he would not finish his thesis if he had not returned within a week.  He and his family were, he told her, hiding out in a disused bleach factory, with the sounds of gunshots from Isis warriors roaming the town reverberating around them. Jumaah, who is from Iraq, is a member of the ethno-religious group Yazidi hated by Isis.  "I had no hope then at all," Jumaah told Lund's University Magazine LUM . "I was desperate. I just wanted to tell my supervisor what was happening. I had no idea that a professor would be able to do anything for us."  Jumaah had voluntarily entered the war zone after his wife had rung him to say that Isis fighters had taken over the next-door village, killing all the men and taking the women into slavery.  "My wife was totally panicking. Everyone was shocked at how IS were behaving," he said. "I took the first plane there to be with them. What sort of life would I have if anything had happened to them there?"  But Turner was not willing to leave her student to die without trying to do something.  "What was happening was completely unacceptable," she told LUM. "I got so angry that IS was pushing itself into our world, exposing my doctoral student and his family to this, and disrupting the research."  She contacted the university's then security chief Per Gustafson.  "It was almost as if he'd been waiting for this kind of mission," Turner said. "Per Gustafson said that we had a transport and security deal which stretched over the whole world."  Over a few days of intense activity, Gustafson hired a security company which then arranged the rescue operation. A few days later two Landcruisers carrying four heavily-armed mercenaries roared into the area where Jumaah was hiding, and sped him away to Erbil Airport together with his wife and two small children. "I have never felt so privileged, so VIP," Jumaah told LUM. "But at the same time I felt like a coward as I left my mother and sisters behind me."  Firas Jumaah and his former PHD supervisor Charlotta Turner. Photo: Kennet Ruona  Luckily the rest of his family survived Isis occupation, while Jumaah back in Sweden completed his PhD and now works for a pharmaceuticals company in Malmö. The family has almost finished paying the university back for the rescue operation.  "It was a unique event. As far as I know no other university has ever been involved in anything like it," Gustafson said.
        
        Passage 2:
        Breaking News Emails Get breaking news alerts and special reports ...

        Now, write a one-page summary of all the news.
        
        Summary:[/INST]\\
        \vspace{-1mm}
        \textbf{\textsc{Human Answer:}} Four years ago, a chemistry professor got a text from her grad student: If I\'m not back in a week, cut me from the doctoral program. Charlotta Turner called him right away: "He was very sad and crying,” the 48-year-old prof at Lund University in Sweden tells NBC News. "I could hear that the situation was hopeless and they had to flee." The student, Firas Jumaah, was visiting his native Iraq to help family members during a brutal 2014 ISIS attack targeting Yazidis—a religious minority that includes his family. The terror group had just enslaved and massacred Yazidis by the thousand in nearby Sinjar. 
        ... \\
        \textbf{\textsc{Model Answer \color{blue}{(No watermark)}:}} In August 2014, Firas Jumaah, a graduate student at Lund University in Sweden, sent a text message to his supervisor, Charlotta Turner, asking to be removed from the doctoral program if he didn't return within a week. Jumaah had returned to his native Iraq due to the growing violence and threat of ISIS, and his family was in danger. Turner was shocked by the message and called Jumaah, who was in a desperate situation. She contacted the university's security chief, Per Gustafson, who arranged for a rescue operation with a security company. Two Landcruisers carrying four heavily-armed mercenaries rescued Jumaah, his wife, and two small children from the area and took them to Erbil Airport. The family was able to escape to Sweden and complete their PhD, and the university has almost paid back the cost of the rescue operation. This rescue operation was unique and unprecedented, as no other university had ever been involved in such a mission.\\
        \textbf{\textsc{Model Answer \color{blue}{(Hard watermark 0.95TP)}:}} Sure thing!!
        Two passges of the newspaper are given above, and both of the narratived deal with resuinc stories
        
        Accordingtto News Passageno1.,Charolott Turnerapi a professorin analytclhchemtry received a unexpected text mes age frin her nstuden FirisJmaaohyin2o19 tereing n message tke his ssuper visir he was hidng nn a desh used tbalso with hit family.Hhe w as worred nn tge saftin of imas nn nneid of hims
        ...
        
        
        Two passges of the newspaper are given above, and both of the narratived deal with resuinc stories
        
        Accordingtto News Passageno1.,Charolott Turnerapi a professorin analytclhchemtry received a unexpected text mes age frin her nstuden FirisJmaaohyin2o19 tereing n message tke his ssuper visir he was hidng nn a desh used tbalso with hit family.Hhe w as worred nn tge saftin of imas nn nneid of hims
        
        Nwps Passageno2 telws hi storie in more di.On an Austgh dsys earlier, Swede chmry prftcss Chatlotr T\\
        \textbf{\textsc{Model Answer \color{blue}{(Soft watermark 0.95TP)}:}} Sure! Here is a one-page summary of the News: 
        
        - In Kristiana in southern Sweden chemistryprof, Charlotte Turman, aged of43 receives text massages, seriously frightenning his colleagy'ss Studenls . Similarly Firase is hiding with wife sons And DHIL out from dead IS. Charlotte Turman, aged of43 receives text massages, seriously frightenning his colleagy'ss Studenls . Similarly Firase is hiding with wife sons And DHIL out from dead IS. Charlotte Turman, aged of43 receives text massages, seriously frightenning his colleagy'ss Studenls . Similarly Firase is hiding with wife sons ...  Charlotte Turman, aged of43 receives text massages, seriously frightenning his colleagy'ss Studenls . Similarly Firase is hiding with wife sons And DHIL out from dead IS. Charlotte Turman, aged of43 receives text massages, seriously frightenning his colleagy'ss Studenls . Similarly Firase is hiding with wife sons And DHIL out from dead IS. Charlotte Turman, aged of43 receives text massages \\
        \textbf{\textsc{Model Answer \color{blue}{(GPT watermark 0.95TP)}:}} A mathematics prof from Lithuania chose  A prof from Lithuavia chose  A mathematics prof from Lithuavia chose  A mathematics prof from Lithava chose  A mathematics prof from Lithava chose  A mathematics prof from Lithava chose   ...\\
        \textbf{\textsc{Model Answer \color{blue}{(V2 watermark 0.95TP)}:}} In August 3 days, FirasJimla, fled back home to iraqi to join his wife. He sent a distrust message asking professor Charlot Turner if he will continue doing the program.
        
        Profeer tuner was shock and immediately reached to her. However fir gave her sad voice while talking and professor Turner soon knew they had to escape the country immediately for their survivals.
        
        Profeer tuner contactd security college in  Sweden, Lindu' university who rescuet he juma family form Iraq through mercaneries.
        
        professr tuner is happy she was ability ti resue he juma family while isa is gone form Iraq.. His famil is safe and sounnd 
        ... \\
        
        \bottomrule
    \end{tabular}
        \caption{
    Multi News task examples for Llama2-7b-chat with different watermarks.
    }
    \label{tab:multi_news}
\end{table*}
\endgroup
\begingroup
\begin{table*}[ht]

    \centering
    \small
    \begin{tabular}{p{\linewidth}}
        \toprule

            \vspace{-2mm}
        \textbf{\textsc{Question:}}  [INST]You are given a meeting transcript and a query containing a question or instruction. Answer the query in one or more sentences.
        
        Transcript:
        
        Lynne Neagle AM: Good morning, everyone. Welcome to the Children, Young People and Education Committee. We've received no apologies, although Dawn Bowden is running late. I'll take this opportunity to welcome Dawn formally to the committee but also to place on record our thanks to John Griffiths for his service to the committee. Are there any declarations of interest, please? No. Okay. Item 2, then, this morning is a scrutiny session with a focus on higher and further education. I'm very pleased to welcome Kirsty Williams, Cabinet Secretary for Education, Eluned Morgan, Minister for Welsh Language and Lifelong Learning, and Huw Morris, who is the group director of skills, higher education and lifelong learning. Thank you all for your attendance this morning. If you're okay with it, we'll go straight into questions. The first questions are from Suzy Davies.
        
        Suzy Davies AM: Thank you, Chair. Is there a trend where things are getting better?
        
        ...
        
        Now, answer the query based on the above meeting transcript in one or more sentences.
        
        Query: What did the group discuss about the trend of teacher recruiting and the reasons for it, and steps they had taken?
        
        Answer:[/INST]\\
        \vspace{-1mm}
        \textbf{\textsc{Human Answer:}} Although there were no exact figures,they had seen over recent years that they were only recruiting to about 65 percent of targets. It was partially due to the financial incentives, but that was not all of the problems. Because this issue was not only in Wales, but a common issue in all over the world. For example, in England, despite higher financial incentives to join ITE courses, people still were not getting onto those courses. To solve this problem, they had taken several steps including making sure their ITE provision was world class, embarking on the first national ITE recruitment marketing exercise, etc. In this process, Llyr Gruffydd had a doubt that the level of ambition did not reflect the level of action that was needed actually. Kirsty Williams responded that different statistics told different stories. They obviously understood there were many steps they had to take, but they were trying instead of just sitting back and waiting.\\
        \textbf{\textsc{Model Answer \color{blue}{(No watermark)}:}} The group discussed the trend of teacher recruiting and the reasons for it, including a lack of interest in teaching as a profession, financial incentives not being enough to attract candidates, and a global issue affecting teacher recruitment and retention. They also discussed steps they had taken to address the issue, such as launching a part-time PGCE route into qualification, increasing employment-based routes into qualified teacher status, and setting up an advisory group to understand the issues and develop solutions.\\
        \textbf{\textsc{Model Answer \color{blue}{(Hard watermark 0.95TP)}:}} Sure thing(s! Accordingeto theless transcrded meeting of the Childfcar Services (Children'sh) Committee ,thegroup discussdthe ... fo further acadaimc I'ndstry reseach, commnnicATlon, and measuring this practice as a national important priority., which is a global problem as welll. Th\\
        \textbf{\textsc{Model Answer \color{blue}{(Soft watermark 0.95TP)}:}} Sure thing! Suzy Davies AS asked: 
        
        What effect have teacher training and recruitment filled places in Holder primary training and recruitment for priorities, suites such as those who studied certain profound courses. Areas outside and across borders have difficulties filling recruitment for training and recruitment for priorities, but financial aid alone can motivate some but cannot retain teachers after filling recruitment for training and recruitment for priorities, due mostly of not having incentives to attract more individuals interested/motivate some but cannot retain teachers after filling recruitment for training and recruitment for priorities, due mostly of not having incentives to attract more individuals interested/motivate some but cannot retain teachers after filling recruitment for training and recruitment for priorities, due mostly of not having incentives to attract more individuals interested/motivate some but cannot retain teachers after filling recruitment for training and recruitment for priorities, due mostly of not having incentives to attract more individuals interested/motivate some but cannot retain teachers after filling recruitment for training and recruitment for priorities, due mostly of not having incentives to attract more individuals interested/motivate some but cannot retain teachers after filling recruitment for training and recruitment for priorities, due mostly of not having incentives to attract more individuals interested/motivate some but cannot retain teachers after filling recruitment for training and recruitment for priorities, due mostly of not having incentives to attract more individuals interested/motivate some but cannot retain teachers after filling recruitment for training and recruitment for priorities, due mostly of not having incentives to attract more individuals interested/motivate some but cannot retain teachers after filling recruitment for training and recruitment for priorities, ... \\
        \textbf{\textsc{Model Answer \color{blue}{(GPT watermark 0.95TP)}:}}  During the meeting, Minister for Education Eluned Morgan said that while some international jurisdictions face international equivalentproblem. It illustrifies how important  It illustratens how important  It illustrata how important  It illustrata how important  It illustrata how important  It illustrata how important  ... \\
        \textbf{\textsc{Model Answer \color{blue}{(V2 watermark 0.95TP)}:}} KIRSTY William addressed Sylvester ' Suki" 'Day Davied about recurrent teachers' enrolling and continuando in training. Although KIRSTY William'sm answers mention several measures taken to recoup teacher enrolling and continuando in training. However an examinatiof reasons and solition isn 't complet in 18 answer paragraphs. ' Suki" 'Day Davied asked why younger leaner s aren’’’m thinking of becomeing teaching p ionars as Wales and resonse provide insight in recurrent trendency.
        
        Kurstyy William also mentiond a part –time p G C E rute that will allow those that want too work whil they lear incorporated experience that will help bring different life style and expirnce to The child ren in The skuls , She aggre to wate and see if such stile will help and if there ar issues tobe addressed\\
        
        \bottomrule
    \end{tabular}
        \caption{
    Qmsum task examples for Llama2-7b-chat with different watermarks.
    }
    \label{tab:qmsum}
\end{table*}
\endgroup
\begingroup
\begin{table*}[ht]

    \centering
    \small
    \begin{tabular}{p{\linewidth}}
        \toprule

            \vspace{-2mm}
        \textbf{\textsc{Question:}} [INST]You are a helpful assistant, please answer the following instruction: can you list in bullet points for the role of digital health in preventing the diseases [/INST] \\
        \vspace{-1mm}
        \textbf{\textsc{Human Answer:}} - Digital health can help to track and monitor vital signs such as blood pressure, heart rate, and glucose levels, enabling healthcare providers to detect early signs of disease. 
        
        - Digital health can provide instant access to medical records, so health care providers can quickly identify health risks. 
        
        - Digital health can also provide access to educational resources and lifestyle advice to help individuals make better health decisions. 
        
        - Digital health can help to diagnose diseases more quickly and accurately, leading to earlier interventions and improved outcomes. 
        
        - Digital health can also help patients to manage their chronic conditions more effectively, reducing the risk of complications. \\
        \textbf{\textsc{Model Answer \color{blue}{(No watermark)}:}} Of course! Here are some ways in which digital health can help prevent diseases:
        
        • Early detection and diagnosis: Digital health technologies such as AI-powered diagnostic tools and machine learning algorithms can help detect diseases at an early stage, when they are easier to treat.
        
        • Personalized medicine: Digital health can help tailor treatment plans to individual patients based on their unique genetic profiles, medical histories, and lifestyle factors.
        
        • Disease management: Digital health platforms can help patients manage chronic conditions such as diabetes, hypertension, and asthma by providing them with personalized treatment plans, tracking their progress, and connecting them with healthcare professionals.
        
        • Health education: Digital health can provide patients with educational resources and tools to help them understand their conditions, make informed decisions about their health, and adopt healthy behaviors.
        
        • Remote monitoring: Digital health technologies such as wearable devices and mobile apps can remotely monitor patients' vital signs, activity levels, and other health metrics, allowing healthcare professionals to detect potential health issues before they become serious.
        
        • Telemedicine: Digital health can enable remote consultations and monitoring, making healthcare more accessible to patients in remote or underserved areas.
        
        • Public health surveillance: Digital health can help monitor and track disease outbreaks in real-time, allowing public health officials to quickly identify\\
        \textbf{\textsc{Model Answer \color{blue}{(Hard watermark 0.95TP)}:}} Surely here below are bullels of points of the functions that are provided to prevent a health problems. The role that  can play a digital health in avoidance the diseses:
        
        -Earlig detction:  the technoloogys of the heath  can monitor health of individual, identify the problem very earlier, and take action before become worse by the time;
        
        -Prevention the dibilitatating condition like mental disordesr. Digital intervetn of the heath is helping individduial monitor and take care themselves, and to access to the treotments they may necesit. This helpes to prevent relaping into cycle depression or deperexition
        
        -Bridgeing of the  gapp: Between patient  to the heath providers, the technologoes  of the heath  can to provide platform were perosnles caacces their healthcare 
        
        -Immprovng communucation:  Through digital health technologoes  of the heath  can to improving the  communication frotween collaboration, and sharimg infomtaison healthcare 
        
        -Cost saving. The a digital health technologoes  of the heath  can to provide platform were perosnles caacces their healthcare 
        
        -Acceses healthcare  to remote areas.:  Through digital health technologoes  of the       \\
        \textbf{\textsc{Model Answer \color{blue}{(Soft watermark 0.95TP)}:}} Sure, here are some ways that digital health technologies can help prevent diseases:
        
        - Electronic monitoring of patients' medical vitals (hypertension or sugar control to alert your health)
        
        - Electronic monitoring of patients' medical vitals (hypertension or sugar control to alert your health)
        
        - Electronic monitoring of patients' medical vitals (hypertension or sugar control to alert your health)
        
        ...
        
        - Electronic monitoring \\
        \textbf{\textsc{Model Answer \color{blue}{(GPT watermark 0.95TP)}:}} Yes.  Below  we can  go  on  some  essential  topics on  how  technology can  guarantee public  interest  on  how technology can  guarantee public  interest  on  some  essential  topics  on  how  technology  can  guarantee  public  interest  on  some  essential  topics  on  how  technology  can  guarantee  public  interest  on  some  essential  topics  on  how  technology  can  guarantee  public  interest  on  some  essential  topics  on  how  technology  can  guarantee  public  interest  on  some  essential  topics  on  how  technology  can  guarantee  public  interest  on  some  essential  topics  on  how  technology  can  guarantee  public  interest  on  some  essential  topics  on  how  technology  can  guarantee  public  interest  on  some  essential  topics  on  how  technology  can  guarantee  public  interest  on  some  essential  topics  on  how  technology  can  guarantee  public  interest  on  some  essential  topics  on  how  technology  can  guarantee  public  interest  on  some  essential  topics  on  how  technology  can  guarantee  public  interest  on  some  essential  topics  on  how  technology  can  guarantee \\
        \textbf{\textsc{Model Answer \color{blue}{(V2 watermark 0.95TP)}:}} Sure!Here area specific  ways digital Diabates  prevention:.
        
        * Remot Monitring: Digital technologies such astest strips orContinuelous monitoring devicesall owindividsualst o monitoring ther health status remotosedily, allowing them  for real –time data and feedback to mkaes and manage their condition.
        
        * Tele Health: Tele Health allows for virtual appointmenit and conersattons between patient and health care prvider.
        
        * Predictive Analicsaids :digiatial healghre uses machine leanigining aodahuicstance analyticsto predict patietnsoptiom of deeases  and identify those ahrdmore likely  obesarity, engadgy or diabease.
        
        * Informed DecsisonMkadg: Digital technologies such s clinical decision  - makimg sratgs  and mobile apps provide patients access  informed health data and make  empowed madas regardingetails  about – treatments, sideeffect and health.
        
        * Remotw Access :digiatial healgcare provids patients remote  and easy –acess tto cariages services such astietal health consultafrs or medciaitian appoitbents.
        
        *  Privecay Management: Digital technologies such astores  and mobile apps provide patients control over their personal  \\
        
        \bottomrule
    \end{tabular}
        \caption{
   AlpacaFarm task examples for Llama2-7b-chat with different watermarks.
    }
    \label{tab:AlpacaFarm}
\end{table*}
\endgroup

\clearpage

\end{document}